\documentclass[sigconf,nonacm]{acmart}
\renewcommand\footnotetextcopyrightpermission[1]{}
\acmConference{}{}{}
\acmBooktitle{}
\usepackage{natbib}
\usepackage{wrapfig}
\usepackage{amsmath}

\usepackage{amssymb}
\usepackage{mathtools}
\usepackage{amsthm}
\usepackage{microtype}
\usepackage{graphicx}
\usepackage{subcaption}
\usepackage{booktabs}
\usepackage{hyperref}
\usepackage{multirow}
\usepackage{arydshln}
\usepackage{todonotes}
\usepackage{soul}


\begin{document}
\title{Reliable OOD Virtual Screening with \\
Extrapolatory Pseudo-Label Matching}

\author{
Yunni Qu \textsuperscript{\rm 1}, Bhargav Vaduri\textsuperscript{\rm 1}, Karthikeya Jatoth\textsuperscript{\rm 1}, James Wellnitz \textsuperscript{\rm 2}, Dzung Dinh\textsuperscript{\rm 1}, Seth Veenbaas\textsuperscript{\rm 2}, Jonathan Chapman\textsuperscript{\rm 2}, Alexander Tropsha \textsuperscript{\rm 2}, Junier Oliva\textsuperscript{\rm 1}
}

\affiliation{
  \textsuperscript{\rm 1} Department of Computer Science, University of North Carolina at Chapel Hill\\
  201 S Columbia St, Chapel Hill, NC 27599 \country{USA} \\
  }
  \affiliation{
    \textsuperscript{\rm 2} Eshelman School of Pharmacy, University of North Carolina at Chapel Hill \\
  301 Pharmacy Lane, Chapel Hill, NC 27599 \country{USA} 
  \
}

\renewcommand{\shortauthors}{Trovato et al.}

\begin{abstract}
Machine learning (ML) models are increasingly deployed for virtual screening in drug discovery, where the goal is to identify novel, chemically diverse scaffolds while minimizing experimental costs. This creates a fundamental challenge: the most valuable discoveries lie in out-of-distribution (OOD) regions beyond the training data, yet ML models often degrade under distribution shift. Standard novelty-rejection strategies ensure reliability within the training domain but limit discovery by rejecting precisely the novel scaffolds most worth finding. Moreover, experimental budgets permit testing only a small fraction of nominated candidates, demanding models that produce reliable confidence estimates. We introduce EXPLOR (Extrapolatory Pseudo-Label Matching for OOD Uncertainty-Based Rejection), a framework that addresses both challenges through extrapolatory pseudo-labeling on latent-space augmentations, requiring only a single labeled training set and no access to unlabeled test compounds, mirroring the realistic conditions of prospective screening campaigns. Through a multi-headed architecture with a novel per-head matching loss, EXPLOR learns to extrapolate to OOD chemical space while producing reliable confidence estimates, with particularly strong performance in high-confidence regions, which is critical for virtual screening where only top-ranked candidates advance to experimental validation. We demonstrate state-of-the-art performance across chemical and tabular benchmarks using different molecular embeddings.
\end{abstract}
\maketitle


\section{Introduction} \label{intro}
\begin{figure}
  \centering
  \includegraphics[width=0.60\linewidth]{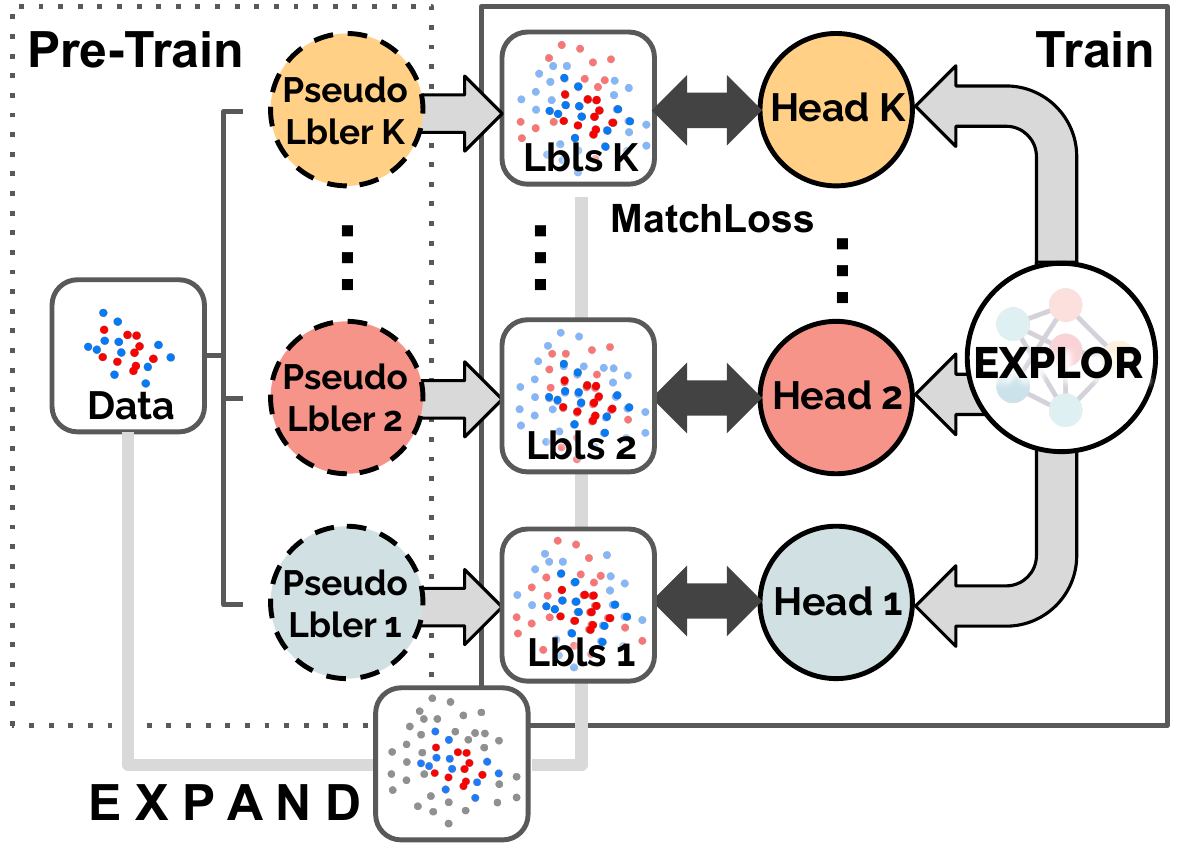} 
  \caption{EXPLOR trains a multi-headed neural network with diverse pseudo-labelers on expanded data.}
    \label{fig:explr_sum}
\end{figure}
Ligand-based virtual screening (LBVS) is a central computational workhorse in modern drug discovery: given a fixed molecular representation (e.g., fingerprints or pretrained embeddings), a model predicts a binary property of interest---such as activity against a target---to triage vast purchasable libraries and nominate a small set of candidates for costly experimental follow-up \citep{Shoichet2004VirtualSO}. Today, purchasable and make-on-demand chemical spaces are firmly \emph{multi-billion} scale: for example, ZINC-22 describes purchasable space growing into the \emph{tens of billions} of molecules \citep{tingle2023zinc}, and the largest make-on-demand catalogs exceed \(\sim\)78 billion enumerated compounds (e.g., Enamine REAL Space).Experimental budgets have not scaled accordingly, resulting in only a fraction of predicted hits being nominated for experiential validation.This operational reality makes LBVS a decision problem dominated by the \emph{top} of the ranked list: the primary objective is not uniformly accurate prediction over all screened molecules, but \emph{high precision among the most confident predicted actives}, where false positives directly consume scarce resources and delay discovery.

Crucially, screening such vast external libraries is also an out-of-distribution (OOD) prediction problem. 
The labeled training set for a given target (or assay endpoint) typically covers a narrow slice of chemical space, enriched around previously explored scaffolds and available measurements.
The most meaningful advances arise from identifying structurally novel chemotypes that lie outside of the explored chemical space in the training domain \citep{hu2017recent}.
Thus, the act of using an LBVS model to search for new chemotypes \emph{should} query the predictor far beyond the support of its training distribution. 
When this extrapolation is poorly controlled, two practical failure modes arise: (i) the model may assign \emph{overconfident} scores to spurious OOD candidates, flooding the top of the ranked list with false positives; or (ii) the model may become overly conservative and reject novelty entirely, undermining discovery. 
Accordingly, an effective LBVS model must extrapolate to novel chemical spaces while remaining reliable in the high-confidence region that determines procurement decisions.

Despite major progress in molecular representation learning and supervised property prediction, LBVS pipelines remain brittle under distribution shift. 
Training data is typically concentrated around well-studied chemotypes and assay regimes, whereas the most valuable discoveries often lie in structurally novel regions of the chemical space. 
As a result, models frequently degrade on out-of-distribution (OOD) compounds \citep{Torralba2011UnbiasedLA, liu2021towards, freiesleben2023beyond}. 
A common safety response in machine learning is novelty-based rejection \citep{Dubuisson1993ASD, hendrickx2024machine}: abstain when an input appears far from the training distribution.
While appropriate for settings where a human can intervene upon rejection, this strategy is fundamentally misaligned with LBVS because it \emph{forbids the extrapolation} needed to discover new chemotypes.\footnote{We use \emph{extrapolation} to denote prediction outside the support of the training distribution.}
In drug discovery, the goal is not to avoid novelty, but to handle it \emph{reliably}: a useful LBVS model must extrapolate beyond the training manifold while still providing confidence estimates that support high-stakes, budgeted selection.

A second, closely related mismatch concerns evaluation. 
Virtual screening has long been recognized as an \emph{early recognition} problem, motivating metrics that emphasize the top of the ranked list such as enrichment factors (EF$_{x\%}$) and BEDROC \citep{truchon2007evaluating, hamza2012ligand}. 
In contrast, OOD generalization work is \emph{commonly summarized with global metrics} (e.g., AUROC or full AUPRC) and calibration summaries (e.g., ECE, NLL, Brier), and many modern molecular uncertainty quantification benchmarks report exactly these quantities \citep{li2023muben}. 
However, global averages can be weakly coupled to screening utility: a method may improve overall discrimination while still producing overconfident false positives among its highest-scoring candidates---the precise failure mode that wastes procurement and assay budget. 
Because real screening campaigns only test a conservative slice of candidates, evaluation should emphasize \emph{early-retrieval reliability}: precision among the most confident predicted actives at low recall. 
To reflect this regime, we evaluate models using a truncated precision--recall area at conservative recall thresholds,
\begin{equation}
\mathrm{AUPRC@R{<}\tau} \;=\; \frac{1}{\tau}\int_{0}^{\tau} \mathrm{Precision}(r)\,dr,
\label{eq:auprc}
\end{equation}
which measures how effectively a model concentrates true positives while minimizing costly false positives \emph{within the top-confidence region} most relevant to LBVS triage.\footnote{We note that partial/truncated PR-area metrics have been used in other evaluation contexts; our focus is their systematic study for OOD reliability in LBVS, where practical decisions are driven by early retrieval rather than global averages \citep{cappellato2022investigating}.}

In addition to statistical challenges, LBVS must operate under strict computational budgets. Although expressive 3D geometric and graph-based pipelines can be effective, their inference cost can be prohibitive for ultra-large screening libraries \citep{alves2022graph, tingle2023zinc, zhou2024artificial}. By contrast, molecular fingerprints and other fixed-length vector representations support much faster screening, making them a practical choice for billion-scale triage. This large-scale, vector-based regime also commonly arises in a data-limited setting: often only a single labeled dataset, concentrated in a narrow region of chemical space, is available for training. We therefore study the practically important and methodologically challenging problem of \emph{single-source} LBVS, where a model trained on one localized source must generalize to structurally novel candidates at test time \citep{Qiao2020LearningTL}.

\textbf{Our approach.}\quad
We introduce \textbf{EX}trapolatory \textbf{P}seudo-\textbf{L}abel Matching for \textbf{O}OD Uncertainty-Based \textbf{R}ejection (\textbf{EXPLOR}), an end-to-end framework designed to bridge the gap between (i) the need to extrapolate to novel chemotypes and (ii) the need for reliable confidence in the high-stakes top-ranked region used for procurement.
EXPLOR combines three established ingredients---latent-space augmentation, diverse pseudo-labeling, and multi-task learning---but makes a key observation: when integrated through a per-head matching objective with a bottlenecked multi-headed architecture (see Fig.~\ref{fig:explr_sum}), these components create emergent behavior suited to single-source OOD generalization.
Concretely, EXPLOR (1) constructs a \emph{diverse} set of pseudo-labelers by training on different feature/instance subsets, (2) generates extrapolative training inputs via latent-space expansion, and (3) trains a multi-headed student to match each pseudo-labeler individually (preserving diversity) while also regularizing the ensemble toward agreement (stabilizing confidence). 
Unlike prior approaches that require modality-specific augmentations \citep{Yun2019CutMixRS} or assume access to multiple domains \citep{Ding2022DomainGB, Jang2023ModalityAgnosticSL, Dou2019DomainGV}, EXPLOR is fundamentally modality-agnostic and operates directly on real-valued vectors, making it compatible with fingerprints and a broad range of learned molecular embeddings.

\textbf{Summary of results.}\quad
Across multiple chemical OOD benchmarks, EXPLOR improves over supervised and semi-supervised baselines in the regimes most relevant to screening, with especially strong gains at conservative recall (AUPRC@R$<\tau$) where procurement decisions are made. 
We further find that EXPLOR yields more stable OOD behavior across training trials and identifies a broader set of high-confidence OOD candidates, consistent with improved extrapolatory coverage rather than narrow memorization.

\textbf{Contributions.}\quad
Our main contributions are:
(1) We emphasize \emph{OOD evaluation} for ligand-based virtual screening under strict experimental budgets, prioritizing early-retrieval reliability (precision among top-confidence predicted actives) to better reflect practical triage.
(2) We propose \textbf{EXPLOR}, a modality-agnostic single-source OOD framework that combines diverse pseudo-labeling with latent-space expansion and multi-headed matching to support controlled extrapolation while maintaining reliable confidence in the top-confidence region.
(3) We provide an application-grounded benchmark against a \emph{diverse suite} of competitive OOD generalization and semi-supervised baselines on chemical distribution shifts, and analyze stability and high-confidence OOD coverage to clarify the practical implications for screening.

\section{Related Work}

\textbf{Ligand-based virtual screening under distribution shift and uncertainty.}\quad
LBVS models are typically trained on historical bioactivity measurements but deployed to triage large external libraries whose chemistry differs from the training distribution.
This makes screening inherently an \emph{OOD prediction} problem: the most valuable candidates are often structurally novel, yet extrapolation can induce overconfident false positives that dominate the purchased set.
Prior work has therefore emphasized the importance of uncertainty quantification for molecular property prediction in discovery workflows where predictions guide expensive experiments \citep{hirschfeld2020uncertainty}, and has connected uncertainty-aware modeling to improved screening outcomes in retrospective settings \citep{soleimany2021evidential}.
EXPLOR builds on this motivation but focuses specifically on \emph{screening-relevant reliability}: trustworthy confidence in the extreme top of the ranked list \emph{under OOD shift}.

\textbf{Molecular featurization for LBVS: fingerprints and pretrained representations.}\quad
Much of modern LBVS operates on fixed molecular featurizations, ranging from expert-crafted fingerprints to pretrained neural embeddings. 
Extended-connectivity (Morgan/ECFP) fingerprints remain a widely used and competitive baseline for molecular property prediction \citep{rogers2010extended}, and recent comparative studies find that expert-based representations can perform strongly---and are often easier to deploy---relative to more computationally demanding task-specific neural representations \citep{stepivsnik2021comprehensive}. 
At the same time, representation learning has produced transferable molecular embeddings that can be reused as plug-in features for downstream binary activity prediction, including SMILES language models such as MoLFormer \citep{ross2022largescalechemicallanguagerepresentations} and ChemBERTa \citep{Chithrananda2020ChemBERTaLS}, and graph/geometry-aware self-supervised pretraining methods such as GROVER \citep{rong2020self}, GraphMVP \citep{Liu2022GraphMVP}, MolCLR \citep{wang2022molecular}, and Uni-Mol \citep{zhou2023uni}. 
These trends motivate EXPLOR's modality-agnostic formulation: rather than assuming proteins, docking poses, or modality-specific augmentations, EXPLOR operates directly on real-valued vectors (fingerprints or embeddings) and targets screening-relevant reliability under OOD shift.


\textbf{Decision-aligned evaluation for screening: early recognition vs.\ global metrics.}\quad
Virtual screening has long been recognized as an \emph{early recognition} problem---only a small top-ranked subset is tested---motivating metrics that emphasize early retrieval such as enrichment factors and BEDROC \citep{truchon2007evaluating, hamza2012ligand}. 
In contrast, much of OOD generalization and molecular uncertainty benchmarking reports global AUROC/AUPRC and aggregate calibration metrics, which can be weakly coupled to the hit rate among the highest-confidence predicted actives. 
While partial/truncated PR-area metrics have been used in other evaluation contexts \citep{cappellato2022investigating}, EXPLOR systematically studies early-retrieval precision (low-recall truncated PR area) under chemical OOD shifts to better reflect the practical regime that determines procurement utility.

\textbf{OOD generalization and robust learning.}\quad
Domain generalization methods often assume multiple labeled source domains and seek invariant predictors through explicit domain labels or multiple environments \citep{Li2018DomainGV, Arjovsky2019InvariantRM, Koyama2020OutofDistributionGW, Ding2022DomainGB}. In contrast, LBVS frequently presents a \emph{single-source} regime: one labeled assay dataset and a large, unlabeled external library. 
EXPLOR targets this setting by expanding the effective training support (latent-space expansion) and constraining the learned representation to satisfy multiple diverse labeling functions (multi-head per-expert matching), aiming to enable controlled extrapolation while avoiding brittle overconfidence.\looseness-1

\textbf{Pseudo-labeling and latent-space augmentation.}\quad
Pseudo-labeling and self-training use model predictions as supervisory signals for unlabeled data \citep{Lee2013PseudoLabelT}, with modern variants improving robustness via consistency regularization, augmentation, and selective acceptance \citep{Xie2019SelfTrainingWN, sohn2020fixmatch}. Separately, latent-space augmentations provide a modality-agnostic way to perturb inputs \citep{Cheung2021MODALSMA}. EXPLOR combines these ideas in a distinct way: it constructs \emph{diverse} pseudo-labelers and preserves this diversity by aligning each expert to a dedicated prediction head (per-head matching), while simultaneously regularizing agreement across heads for stability; importantly, these matching signals are applied not only on the original training set but also on extrapolated latent-space samples, directly targeting OOD reliability in the screening-relevant tail.

\textbf{Selective prediction and reject options.}\quad
Selective classification equips predictors with an abstention mechanism to improve reliability by rejecting uncertain inputs \citep{Fumera2000RejectOW, Geifman2017SelectiveCF, Geifman2019SelectiveNetAD}. 
However, naive novelty rejection is misaligned with LBVS because novelty is precisely the point of screening. EXPLOR is motivated by a different operational requirement: identify a subset of OOD candidates for which the model can make \emph{trustworthy high-confidence positive predictions}, while rejecting unreliable predictions, and evaluate this behavior in the early-retrieval regime most relevant to experimental triage.\looseness-1

\section{Method} \label{sec:method}
EXPLOR tackles OOD LBVS in three stages (Fig.~\ref{fig:explr_sum}): (1) build a diverse ensemble of pseudo-labelers via feature and instance subsampling; (2) expand the effective training support using latent-space perturbations (Sec.~\ref{sec:expaug}); and (3) train a multi-headed predictor in which each head is matched to a different pseudo-labeler (Sec.~\ref{sec:var_reduction}).

We assume a single labeled training set $\mathcal{D}=\{(x_i,y_i)\}_{i=1}^N$ with $(x_i,y_i)\sim \mathcal{P}_{\mathrm{in}}$ drawn \emph{iid} and no domain/environment annotations. 
This reflects typical virtual screening workflows, where models trained in a single assay context must generalize to molecules drawn from heterogeneous external libraries spanning structurally distinct regions of chemical space.
For clarity, we present the binary classification case $y_i\in\{0,1\}$ (e.g., active vs.\ inactive), though the framework extends directly to regression and multi-class prediction. We focus on modality-agnostic real-valued inputs $x_i\in\mathbb{R}^d$, covering molecular fingerprints, pretrained molecular embeddings, and other vector representations common in drug-discovery pipelines.


\subsection{Diverse Psuedo-Labelers} 

Standard empirical risk minimization (ERM), e.g.~cross-entropy training, often yields overconfident, unreliable predictions outside of the training distribution support \citep{nguyen2015deepneuralnetworkseasily, yang2024generalizedoutofdistributiondetectionsurvey}. In contrast, ensembling and bagging can improve robustness by averaging over diverse hypotheses \citep{Jordan1993HierarchicalMO,Eigen2013LearningFR, Arpit2021EnsembleOA, Dietterich2007EnsembleMI, Pagliardini2022AgreeTD,Yao2023ImprovingDG}. Motivated by these observations, EXPLOR trains with a \emph{set of diverse pseudo-labelers} that provide multiple, complementary training targets, reducing reliance on any single brittle predictor.

Specifically, EXPLOR constructs $K$ pseudo-labelers $\{g_k\}_{k=1}^K$, where $g_k:\mathbb{R}^d\rightarrow\{0,1\}$, and uses their outputs as pseudo-labels to supervise a multi-headed network. EXPLOR emphasizes \emph{diversity by construction}: each $g_k$ is trained on a different subsample of instances and a different feature subspace, encouraging specialization to distinct regions and complementary ``views'' of the representation space.
Empirically, XGBoost classifiers \citep{Chen2016XGBoostAS} provided strong base pseudo-labelers; however, we also observed improvements with several alternative pseudo-labeling backbones (Sec.~\ref{sec:ablations}).

\subsection{Expansive Augmentation of Training Data} 
\label{sec:expaug}

\begin{figure}
\vspace{-2pt}
    \centering
    \includegraphics[width=.35\linewidth]{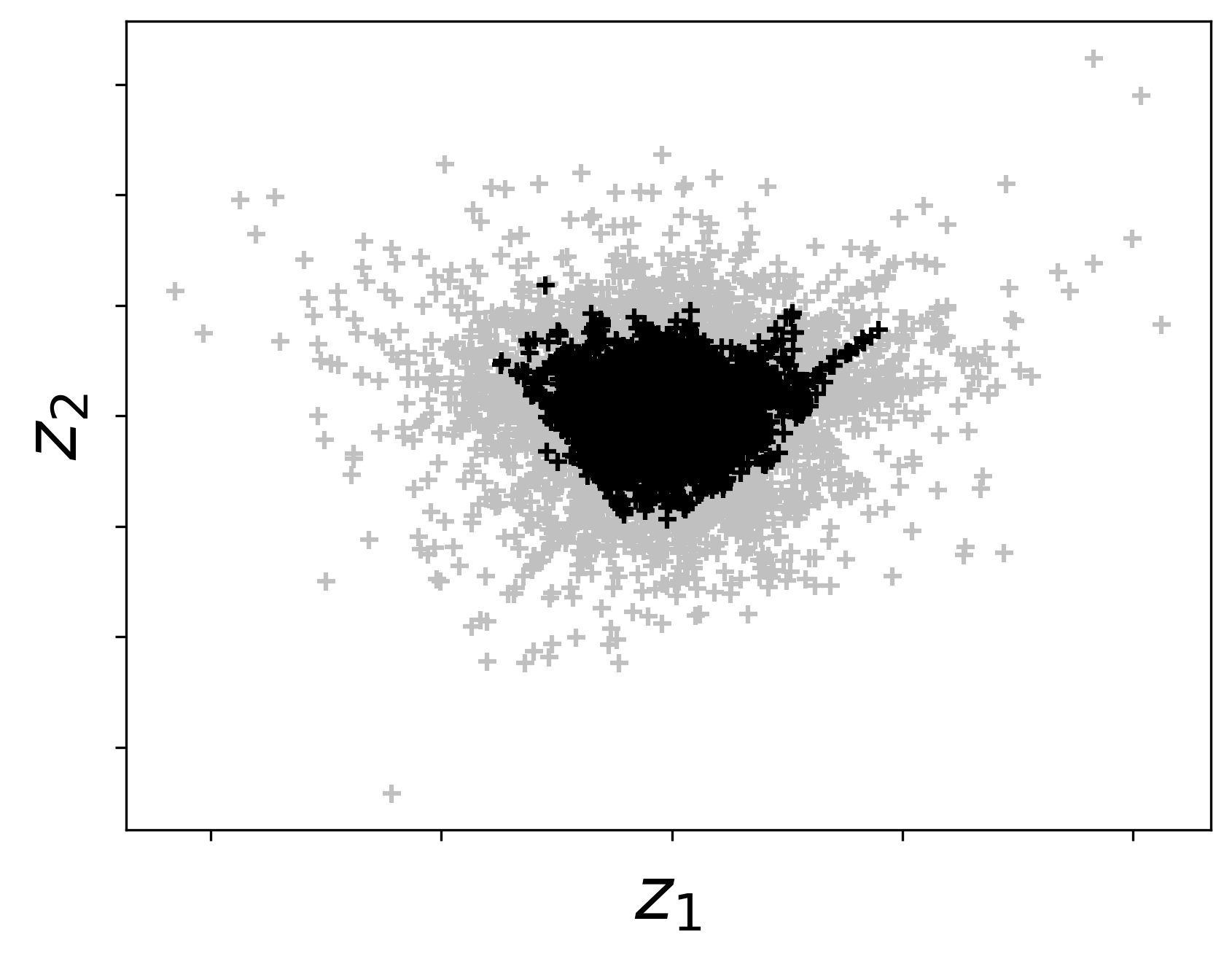}
    \vspace{-2pt}
    \caption{Expansion in latent space: data points (black) are augmented (gray) and expand the distributional support. }
    \label{fig:expand}
\vspace{-10pt}
\end{figure}
To train models capable of extrapolating to OOD samples, we wish to expose them to examples that extend beyond the support of the training distribution. Accordingly, to reason about the support of the training data, and \emph{expand} past it, we propose to leverage a latent factor space, $\varphi: \mathbb{R}^d \mapsto \mathbb{R}^s$. 
While learning semantically meaningful latents is itself an active area, we found autoencoding-based representations effective in practice (see Sec.~\ref{sec:exp}); these provide a decoder $\gamma:\mathbb{R}^s \rightarrow \mathbb{R}^d$. Without loss of generality, we assume the latent space is centered, $\mathbb{E}[\varphi(X)]=0$.

\textbf{Latent expansion.}\quad Given $z=\varphi(x)$, we form an \emph{expansive} perturbation by radial scaling,
\begin{equation}
    z' = (1+|\epsilon|)\,z, \qquad \epsilon \sim \mathcal{N}(0,\sigma^2),
\end{equation}
where $|\epsilon|$ follows a Half-Normal distribution, and decode $x'=\gamma(z')$. This defines a \emph{stochastic} expansion operator on a set of points:
\begin{equation}
    \textbf{Ex}(\{x_i\}_{i=1}^N) \equiv 
    \left\{ \gamma\!\left((1+|\epsilon_i|)\,\varphi(x_i)\right) \,\middle|\, \epsilon_i \sim \mathcal{N}(0,\sigma^2) \right\}_{i=1}^N.
\end{equation}
As illustrated in Fig.~\ref{fig:expand}, $\textbf{Ex}$ pushes samples beyond the in-distribution support and can overlap regions populated by OOD molecules. Unlike small, label-preserving jitters, however, these expansive augmentations do not admit a reliable ground-truth label. We therefore supervise expanded samples via pseudo-labels from our $K$ pseudo-labelers, $\bigl(g_1(x'),\ldots,g_K(x')\bigr)$.\footnote{Alternatively, pseudo-labelers can be applied directly in latent space, $\bigl(g_1(z'),\ldots,g_K(z')\bigr)$, avoiding decoding.}

\subsection{EXPLOR:  Extrapolatory Pseudo-Label Matching for OOD Rejection} \label{sec:var_reduction}

Once latent expansion produces extrapolated inputs, the central challenge is supervision: true labels are unavailable for these OOD samples. We therefore pseudo-label expanded points using our $K$ pseudo-labelers and train a multi-headed network to match them. Concretely, the model comprises a shared MLP embedding $\phi:\mathbb{R}^d \to \mathbb{R}^m$ and $K$ labeler-specific heads $\{h_j\}_{j=1}^K$ (e.g., $h_j:\mathbb{R}^m \to \mathbb{R}$ for binary logits), where head $h_j$ is trained based on pseudo-labeler $g_j$. By requiring a single shared representation to support multiple diverse labeling functions through simple heads, this bottleneck promotes features $\phi$ that are robust across different views of the data.
Our objective has three terms:
\begin{enumerate}
    \item \textbf{Per-head matching} (Eq.~\eqref{eq:head_match}): each $h_j$ mimics its assigned pseudo-labeler $g_j$.
    \item \textbf{Mean matching} (Eq.~\eqref{eq:mean_match}): the ensemble mean prediction matches the mean pseudo-label for consistency.
    \item \textbf{Expanded-data matching} (Eq.~\eqref{eq:expand_match}): apply the same losses on expanded samples.
\end{enumerate}

Our per-head matching loss on a set $\mathcal{S}$ is
\begin{align}    
 &\mathcal{L}_\mathrm{match}(\phi, \{h_j\}_{j=1}^{K}, \{g_j\}_{j=1}^{K}; \mathcal{S})  
  \equiv  \frac{1}{|\mathcal{S}|K} \sum_{x \in S} \sum_{j =1}^K \ell( h_j(\phi(x)), g_j(x)), 
  \label{eq:head_match}
\end{align}
where $\ell(\hat{y},y)$ is a supervised loss (e.g., cross-entropy). We further impose \emph{mean matching} via an $\ell_1$ penalty,
 \begin{align}
      &\mathcal{L}_\mathrm{mean}(\phi, \{h_j\}_{j=1}^{K}, \{g_j\}_{j=1}^{K}; \mathcal{S}) \equiv \nonumber\\ 
& \qquad   \frac{1}{|\mathcal{S}|} \sum_{x \in S}\left \| \frac{1}{K} \sum_{j =1}^K \sigma(h_j(\phi(x))) -  
  \frac{1}{K} \sum_{j =1}^K g_j(x))\right\|_1,
    \label{eq:mean_match}
 \end{align}
with $\sigma(\cdot)$ the sigmoid. The full EXPLOR objective is
\begin{align}
  &\mathcal{L}_\mathrm{EXPLOR}(\phi, \{h_j\}_{j=1}^{K}, \{g_j\}_{j=1}^{K}; \mathcal{D}) \equiv \label{eq:emoeloss}\\  
   &\qquad\   \ \mathcal{L}_\mathrm{mean}(\phi, \{h_j\}_{j=1}^{K}, \{g_j\}_{j=1}^{K}; \mathcal{D}) \label{eq:emoematch}\\
   &\quad \ +\mathcal{L}_\mathrm{match}(\phi, \{h_j\}_{j=1}^{K}, \{g_j\}_{j=1}^{K}; \mathcal{D}) \\
   &\quad \ + \lambda\, \mathcal{L}_\mathrm{match}(\phi, \{h_j\}_{j=1}^{K}, \{g_j\}_{j=1}^{K}; \textbf{Ex}(\mathcal{D})).
   \label{eq:expand_match}
\end{align}
The mean-matching term provides additional supervision on the unexpanded data $\mathcal{D}$. Empirically ($\S$~\ref{sec:exp}), the learned heads often approach the pseudo-labelers' accuracy, and combining both yields the most reliable performance; we therefore deploy the bagged predictor:
\begin{equation}
    f_\mathrm{EXPLOR}(x) \equiv \frac{1}{2K} \sum_{j=1}^K (g_j(x) + h_j(\phi(x))).
    \label{eq:EXPLORpred}
\end{equation}


\subsection{Diversity and Variance Reduction}
We provide intuition for why EXPLOR improves OOD estimation through (i) \emph{diversity-driven multi-task learning} and (ii) \emph{variance reduction/regularization}.

\noindent\textbf{Diversity and multi-task learning.}\quad EXPLOR uses simple heads, which forces the shared encoder to learn a representation $\phi(x)$ from which each pseudo-labeling function can be recovered by a lightweight projection. Formally, the per-head matching loss in Eq.~\eqref{eq:head_match} can be written as a multi-task objective over $K$ \emph{virtual environments}
$\mathcal{E}_j(\mathcal{S})=\{(x,g_j(x))\,|\,x\in\mathcal{S}\}$:
\[
\mathcal{L}_\mathrm{match}(\phi,\{h_j\}_{j=1}^{K},\{g_j\}_{j=1}^{K};\mathcal{S})
= \frac{1}{K}\sum_{j=1}^{K}\mathcal{L}\!\left(h_j(\phi(\cdot)),\,\mathcal{E}_j(\mathcal{S})\right),
\]
where $\mathcal{L}(\cdot,\mathcal{E}_j(\mathcal{S}))$ denotes supervised loss on environment $\mathcal{E}_j(\mathcal{S})$. Training on expanded samples $\textbf{Ex}(\mathcal{D})$ therefore provides supervision (a) \emph{on extrapolated inputs} and (b) \emph{across diverse labeling views} induced by pseudo-labelers trained on different instance/feature subspaces.

\noindent\textbf{Variance reduction and regularization.}\quad Prior work decomposes OOD risk into bias and variance terms \citep{yang2020rethinking,arpit2022ensemble}:
\begin{align}
&\mathbb{E}_{(x,y)\sim \mathcal{P}_\mathrm{out}}\mathbb{E}_{\mathcal{D}\sim \mathcal{P}_\mathrm{in}}
\!\left[\mathrm{CE}\!\left(y,f(x;\mathcal{D})\right)\right]
= \nonumber\\
&\qquad
\mathbb{E}_{(x,y)}\!\left[\mathrm{CE}\!\left(y,\bar{f}(x)\right)\right]
+
\mathbb{E}_{x,\mathcal{D}}\!\left[\mathrm{KL}\!\left(\bar{f}(x),f(x;\mathcal{D})\right)\right],
\end{align}
where $\bar{f}(x)=\mathbb{E}_{\mathcal{D}}[f(x;\mathcal{D})]$. Letting $\bar{g}(x)=\frac{1}{K}\sum_{j=1}^{K}g_j(x)$, we view $\bar{g}$ as a bootstrap-style approximation to $\bar{f}$, yielding the heuristic approximation
\begin{align}
&\mathbb{E}_{(x,y)\sim \mathcal{P}_\mathrm{out}}\mathbb{E}_{\mathcal{D}\sim \mathcal{P}_\mathrm{in}}
\!\left[\mathrm{CE}\!\left(y,f(x;\mathcal{D})\right)\right]
\approx \nonumber\\
& \qquad
\mathbb{E}_{(x,y)}\!\left[\mathrm{CE}\!\left(y,\bar{f}(x)\right)\right]
+
\mathbb{E}_{x,\mathcal{D}}\!\left[\mathrm{KL}\!\left(\bar{g}(x),f(x;\mathcal{D})\right)\right].
\end{align}
This connects to our objective by interpreting expanded samples $\textbf{Ex}(\mathcal{D})$ as a proxy for $\mathcal{P}_\mathrm{out}$ and the matching loss on expansions,
$\mathcal{L}_\mathrm{match}(\phi,\{h_j\},\{g_j\};\textbf{Ex}(\mathcal{D}))$,
as a surrogate for the variance term $\mathbb{E}_{x,\mathcal{D}}[\mathrm{KL}(\bar{g}(x),f(x;\mathcal{D}))]$. Empirically, bootstrap-style trials confirm substantially lower predictive variance on OOD points for EXPLOR relative to ERM (Fig.~\ref{fig:boots}, Tab.~\ref{tab:variance_reduction}).

\section{Experiments} \label{experiments}\label{sec:exp}

\subsection{Datasets}
We curated four bioactivity datasets from ChEMBL \citep{Gaulton2011ChEMBLAL} and Therapeutics Data Commons (TDC) \citep{huang2021therapeutics}: inhibition of human ether-a-go-go-Related Gene (hERG), cytotoxicity in human A549 cells (A549 Cells), agonists for Cytochrome P450 2D6 (CYP2D6), and Ames mutagenicity (Ames) datasets. 
We generated ID and OOD splits using Bemis–Murcko scaffolds, which reduce each molecule to its core ring systems and framework by removing peripheral substituents. Molecules that share the same scaffold were assigned to the same split, ensuring that OOD compounds contain core structural patterns not observed during training. 
This scaffold-based partitioning evaluates whether a model can generalize beyond previously seen structural frameworks, approximating the drug discovery challenge of identifying active compounds with different chemical scaffolds.

We also incorporated the three most challenging OOD gap ligand-based affinity prediction (LBAP) datasets from DrugOOD \citep{ji2022drugood},   ``core EC50,'' ``refined EC50,'' and ``core IC50''.  
Here, `core' and `refined' refer to DrugOOD noise levels, where `core' is the smallest and least noisy dataset followed by `refined' and `general' (not used in this work). EC50 and IC50 are both measures of drug potency, where lower values indicate greater potency.  
Domain splits for the LBAP datasets are defined by molecular size, determined based on the number of atoms in a molecule. We assigned smaller molecules to the ID training set and larger molecules to an OOD validation and OOD testing sets in sequential order. 

Molecules were represented using 1024-bit extended-connectivity fingerprints \citep{ecfp} with radius 3 (ECFP6). Fingerprints are a widely used vectorized chemical representation developed for structure-activity modeling. We selected ECFP6 fingerprints because they provide a strong and computationally efficient molecular representation that has been shown to perform competitively with learned graph-based representations on classification tasks \citep{fp_vs_gcn}.

\subsection{Baselines and Models}
We contextualize our results by comparing against a diverse set of strong OOD generalization methods spanning several complementary approaches. Our code will be open-sourced upon publication.\looseness-1

\subsubsection{\textbf{Baselines}}
In our experiments, baselines span three categories: data augmentation—AdvStyle (adversarial augmentation), Mixup (linear ID sample combinations), and NCDG (augmentation with neuron coverage optimization); ensemble methods—D-BAT (enforcing prediction diversity on OOD data) and EoA (ensembling moving average models); and loss-based robustness—SAM (seeking uniformly low-loss parameter neighborhoods) and UDIM (adversarial latent perturbations with flat minima).
We also compare against semi-supervised methods for tabular data, DivDis and FixMatch, which utilize unlabeled OOD data from the test-time distribution—\emph{more information than EXPLOR}—and thus break from our single-source generalization setting. DivDis trains diverse model heads on unlabeled OOD data, then selects the best with minimal supervision. FixMatch generates pseudo-labels from weakly-augmented OOD samples and trains the model on strongly-augmented versions of the same samples.

\subsubsection{Models}

We considered two EXPLOR models using different pseudo-labelers: one using 1024 XGB Classifiers \citep{Chen2016XGBoostAS} fit to random subsets of data instances and features, and another using a complementary neural model consisting of 64 D-BAT \citep{Pagliardini2022AgreeTD} networks trained in the same way to provide diverse neural pseudo-labels.
For a fair and realistic evaluation, we did not tune EXPLOR-specific hyperparameters (e.g., the number of training iterations or \(\lambda\) in Eq.~\ref{eq:emoeloss}). Instead, we used a fixed architecture: a two-layer MLP with 512 ELU \citep{Clevert2015FastAA} hidden units and linear output heads, with 1024 heads for the XGBoost pseudo-labelers and 64 heads for the D-BAT pseudo-labelers. The latent space was constructed using PCA with 128 components. Additional implementation details are provided in Appx.~\ref{emoe_hp} and Appx.~\ref{sec:appdeats}, including training-time comparisons, where EXPLOR is comparable to, or faster than, other deep learning approaches such as D-BAT, EoA, and AdvStyle.\looseness-1
\begin{figure}[t]
  \centering
  
  \includegraphics[width=0.8\linewidth]{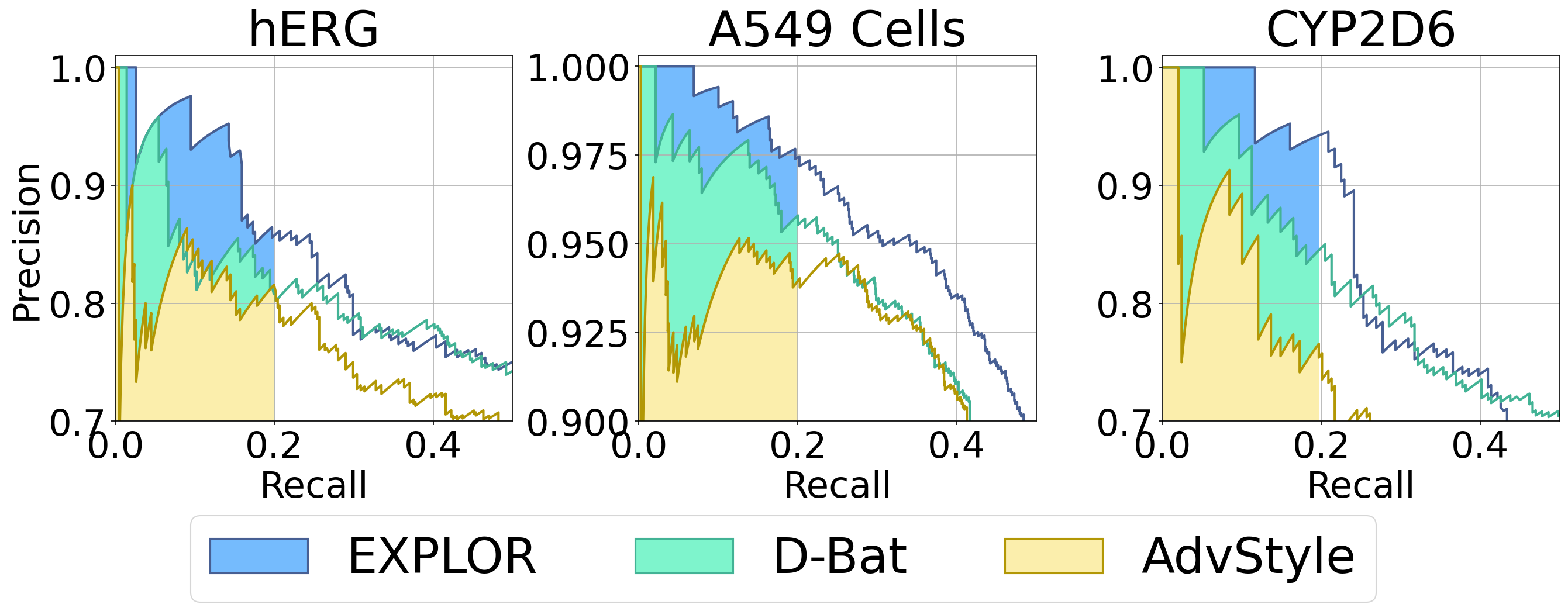} 
  \scriptsize
  \caption{Area under precision/recall curves at recall less than 0.2 (AUPRC@R$<$0.2) for EXPLOR and competitive baselines are shaded below respective precision/recal curves.}
    \label{fig:AUPRC_shaded}
\end{figure}

\subsection{Results}

\begin{figure}[t]
    \centering
    \includegraphics[width=\linewidth]{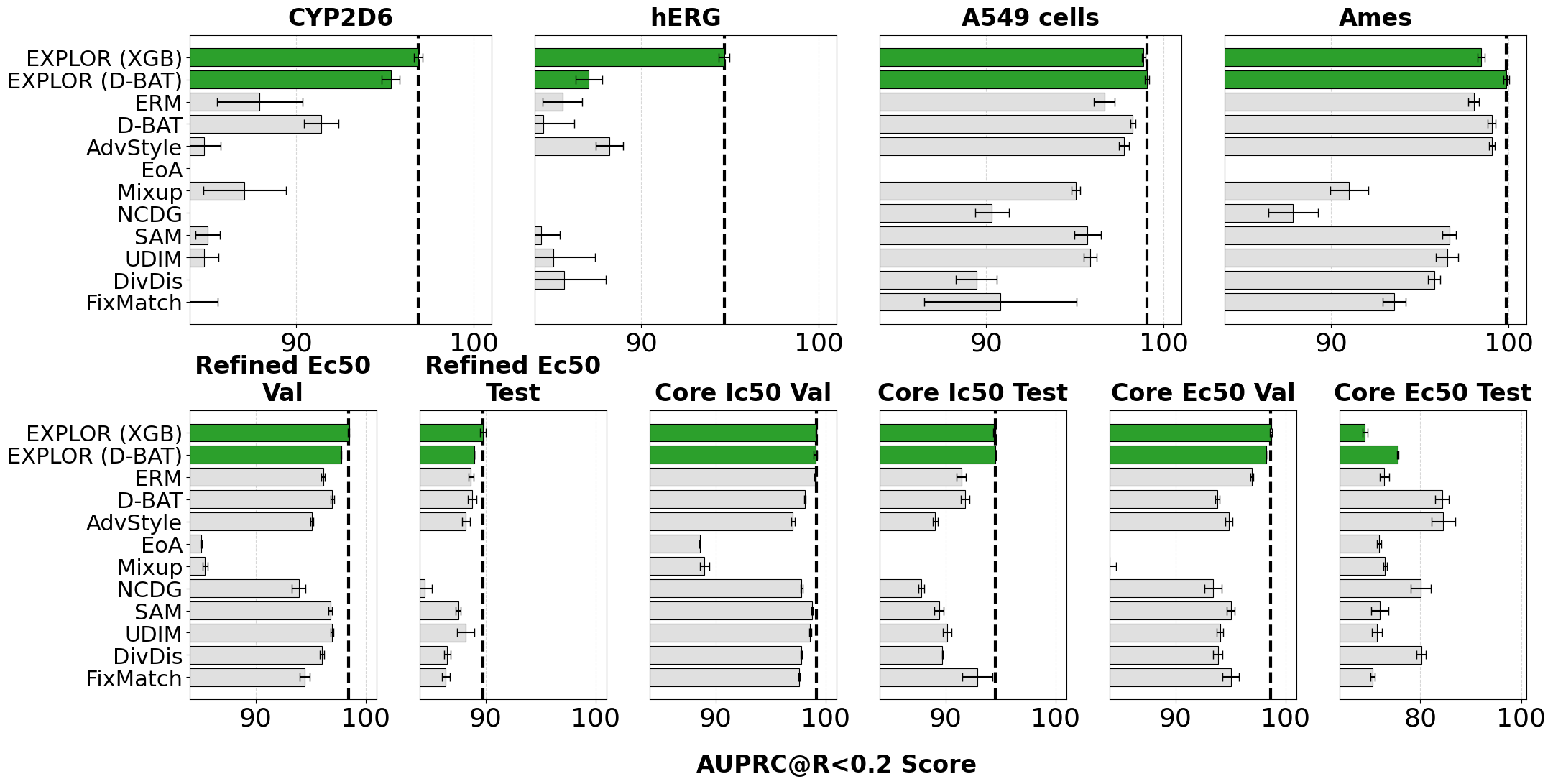}
    \\ \vspace{10pt}
    \includegraphics[width=\linewidth]{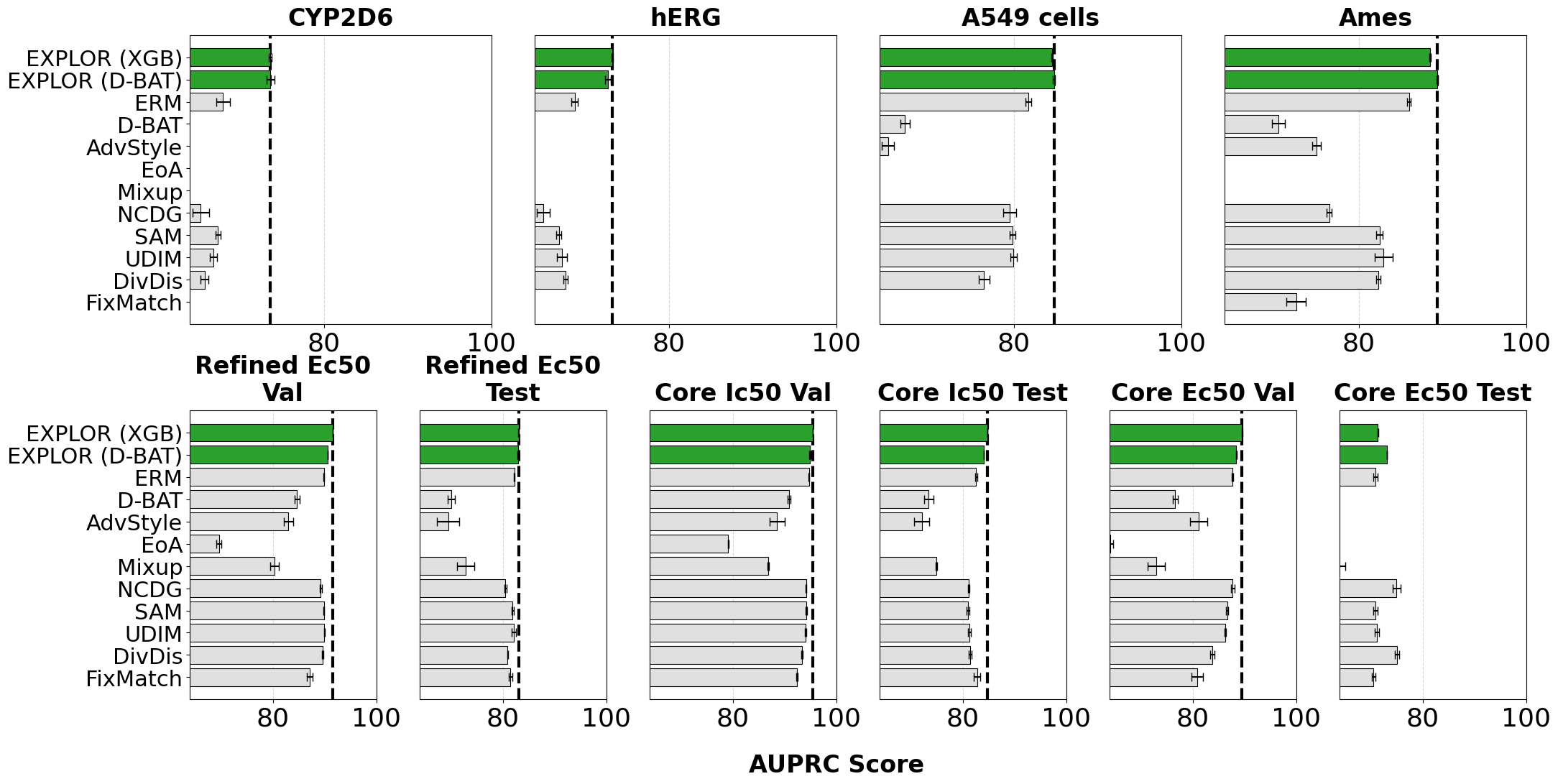}
    \caption{OOD screening performance on scaffold- and size-shifted molecular benchmarks. Datasets where EXPLOR leads are indicated with a vertical dashed line.
Top: early-recall performance measured by $\mathrm{AUPRC}@R{<}0.2$, which emphasizes high-confidence retrieval relevant to practical virtual screening. Bottom: full AUPRC. Higher is better. Error bars denote std.~error across runs.}
    \label{fig:prc}
\end{figure}


\textbf{Metrics} \quad For prediction thresholding (rejection), we directly utilize the conditional probability $P(Y=1 \mid X=x)$ generated by the models.
Because many practical applications such as virtual screening prioritize only the highest-confidence predictions, we emphasize performance in the early-recall regime. In particular, we report $\mathrm{AUPRC}@R{<}\tau$, the area under the precision--recall curve truncated to recall levels below $\tau$ (see Eq.~\ref{eq:auprc} for the formal definition and Fig.~\ref{fig:AUPRC_shaded} for an illustration). We also report full AUPRC and AUROC.
Note that for a fixed screening library, the enrichment factor at a given cutoff is proportional to precision at that cutoff:
\[
\mathrm{EF} = \frac{\mathrm{Precision}}{\pi_{+}},
\]
where $\pi_{+}=A/N$ denotes the overall active prevalence in the candidate set. Hence, EF and precision induce the same ranking of methods on the same dataset, although EF rescales precision by the baseline active rate. Moreover, since $\mathrm{AUPRC@R}\!\!<\!\!\tau$ measures the average precision among the model's most confident positive predictions up to $\tau\%$ recall, dividing by the dataset prevalence yields the average enrichment factor over that same early-recall regime.

\begin{figure}[b]
    \centering
    \includegraphics[width=\linewidth]{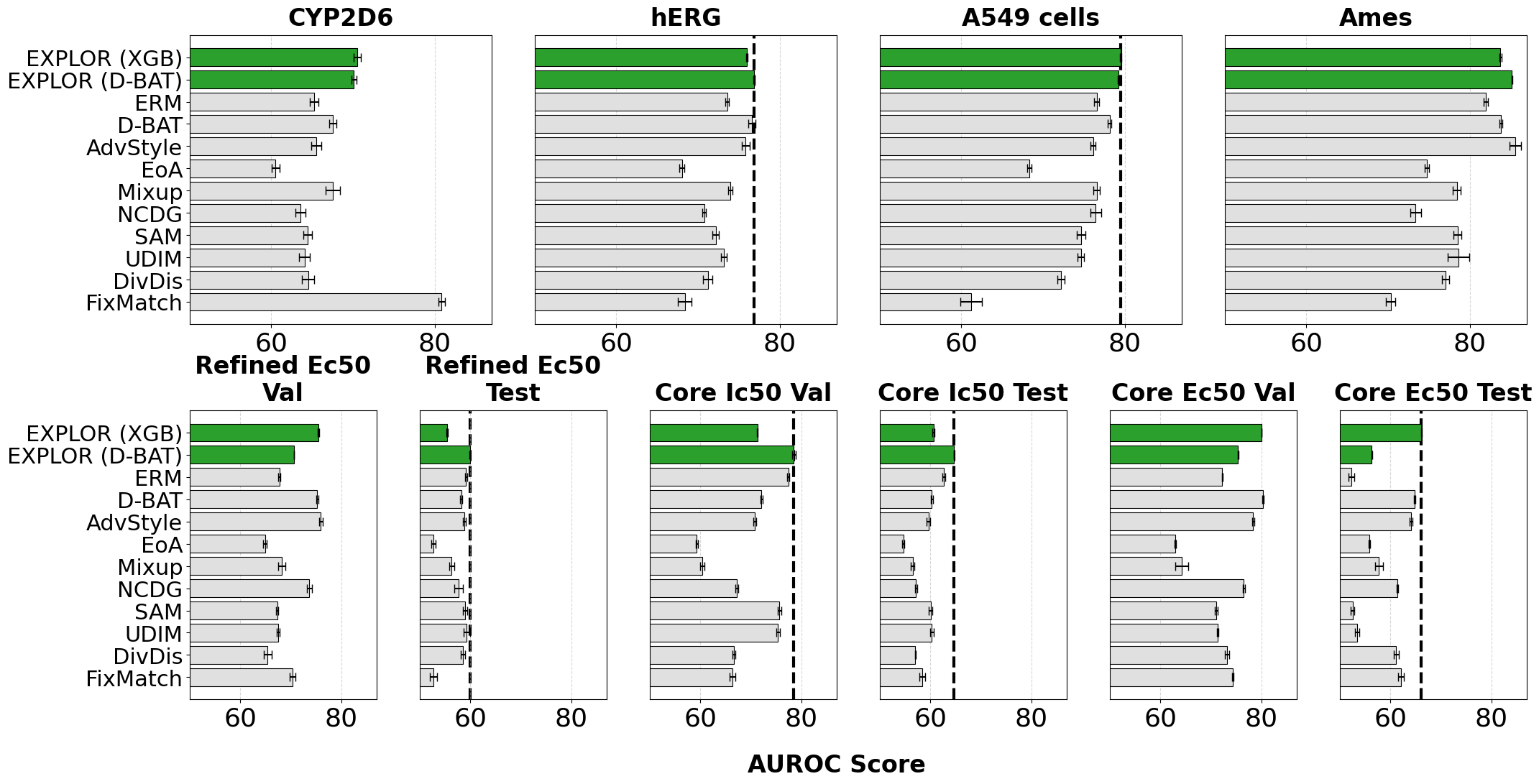}
    \caption{\textbf{AUROC on OOD molecular screening benchmarks.} Higher is better.  Error bars denote std.~error across runs.}
    \label{fig:roc}
\end{figure}

\textbf{Discussion}\quad 
EXPLOR shows advantages in the high-confidence OOD screening regime, while remaining competitive---and often still leading---on full AUPRC and AUROC (Fig.~\ref{fig:prc}, Fig.~\ref{fig:roc}). On the scaffold-based bioactivity benchmarks (CYP2D6, hERG, A549 Cells, and Ames), both EXPLOR variants are typically the best or among the best in the early-recall $\mathrm{AUPRC}@R{<}0.2$ regime, which is the most operationally relevant setting for ligand-based screening because only a small fraction of top-ranked compounds can be advanced to experimental follow-up. 
Full AUPRC/AUROC provide complementary evidence that these improvements are not limited to a narrow operating point: EXPLOR remains at or near the top across scaffold-based benchmarks, indicating strong global discrimination under structural shift even when the largest gains occur in early-recall triage.

A similar pattern holds on the DrugOOD ligand-based affinity prediction tasks. Across refined EC50, core IC50, and core EC50, EXPLOR is frequently among the strongest methods on validation and test splits and often achieves the best or near-best performance under $\mathrm{AUPRC}@R{<}0.2$, AUPRC, and AUROC. These gains are obtained in the single-source setting, unlike DivDis and FixMatch, which additionally leverage unlabeled target-domain OOD samples. Across metrics, the XGB-based pseudo-labeler is often slightly stronger than the D-BAT-based variant, though both generally outperform or match the baselines, suggesting that the main benefit comes from EXPLOR's pseudo-label-matching framework rather than from any single pseudo-labeler family. (See Sec.~\ref{sec:varypl} for additional analysis across different pseudo-labelers.) Overall, the metric profile is consistent: EXPLOR maintains strong global discrimination under OOD shift while delivering its largest practical benefit in reliable early-recall enrichment among top-ranked screening candidates.\looseness-1

\subsection{Results Analysis}

\subsubsection{\textbf{Stability Analysis}}
A key motivation behind EXPLOR is to reduce variance across model fits from the same data source, which, as discussed in Sec.~\ref{sec:var_reduction}, is important for OOD generalization. In practical LBVS settings, standard neural networks can produce overconfident predictions \cite{wei2022mitigating}. When combined with fit instability, this means that even small perturbations to the training data can cause large changes in the evaluation of screened compounds. As a result, unstable overconfident models may nominate markedly different top-ranked screening candidates, undermining screening reliability.\looseness-1

\begin{figure}[b]
        \centering
    \includegraphics[width=0.78\linewidth]{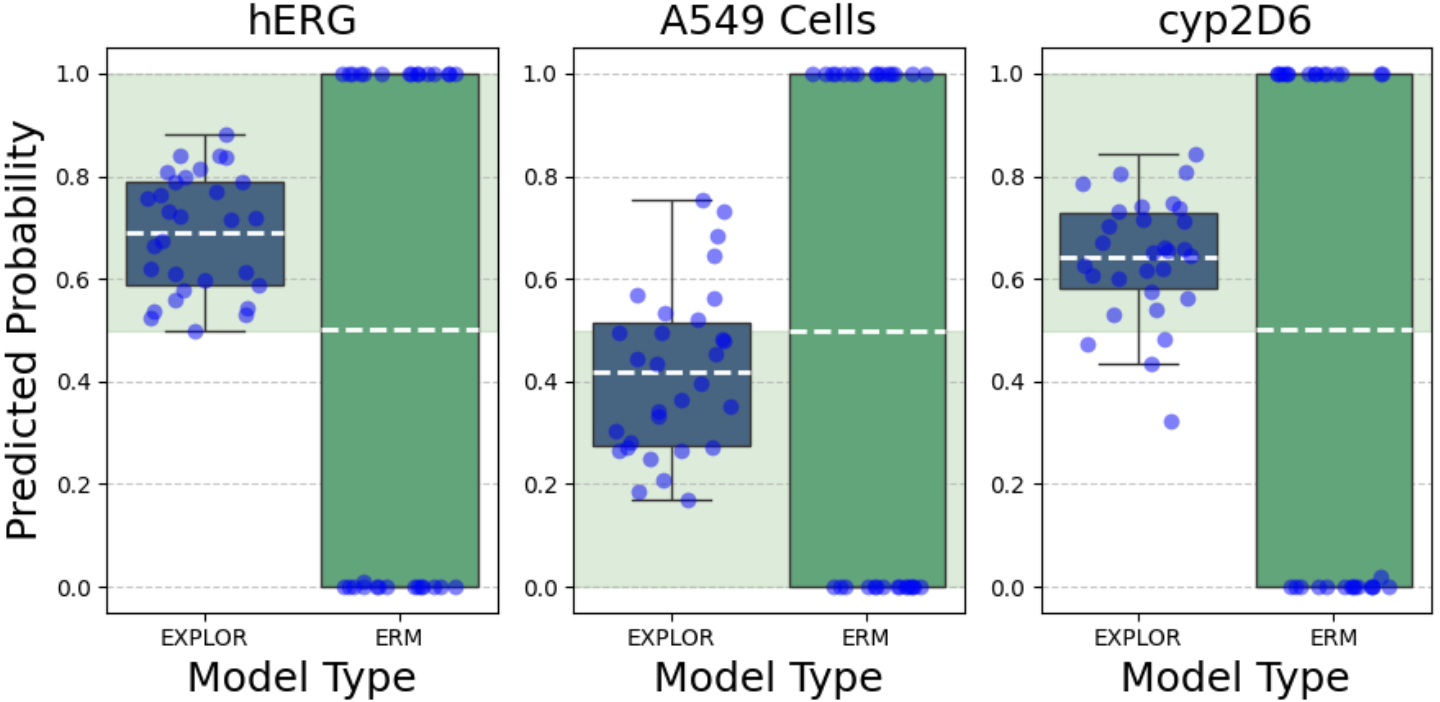}

    \caption{Box plots showing predicted probabilities for EXPLOR's three highest-variance OOD test instances across ChEMBL datasets, comparing EXPLOR vs. ERM. Each scatter point represents the prediction across one of the 30 models trained on bootstrapped samples of the training data (with x-axis noise added to prevent occlusions). The green shaded background indicates region of correct prediction for the respective positive, negative, and positive shown OOD instance (across hERG, A549 Cells, and CYP2D6, respectively). }
    \label{fig:boots}

\end{figure}

We study the stability of EXPLOR versus empirical risk minimization (ERM) training of a neural network across 30 bootstrap-type trials (each using a random subsample the in-distribution training data) and evaluated the variance of predicted probabilities on the same OOD held-out compounds from respective datasets. From each dataset, we plot the \emph{most challenging} (i.e. least stable) instances for EXPLOR in Fig.~\ref{fig:boots}. We note that although these are \emph{the worst-case variance instances} for EXPLOR, EXPLOR still, on average, correctly labels these instances. Moreover, we see that the ERM trained models equally oscillate between positive and negative predictions (at maximum confidence), resulting in the highest possible variance for these examples. 
In these difficult examples, ERM-trained networks exhibit the unreliable screening behavior described above: small perturbations to the training data across bootstrap trials cause the model to oscillate between confidently including and confidently excluding the same compounds as screening candidates.

\begin{table}
\vspace{0pt}
\caption{Predictive variance across 30 bootstrap-trained models on OOD Data. 
Lower values indicate more stable predictions under distribution shift.} 

\label{tab:variance_reduction}
\resizebox{0.7\columnwidth}{!}{

\begin{tabular}{lccc}
\toprule
\textbf{Method} & \textbf{hERG} & \textbf{A549 Cells} & \textbf{CYP2D6} \\
\midrule
ERM & 0.117 & 0.119 & 0.156 \\
EXPLOR Neural Network & 0.023 & 0.028 & 0.025 \\
EXPLOR & \textbf{0.015} & \textbf{0.019} & \textbf{0.016} \\
\bottomrule
\end{tabular}
}
\end{table}

These observations also hold at an aggregate level. Tab.~\ref{tab:variance_reduction} reports the mean predictive variance across OOD held-out samples for ChEMBL datasets. One can see that the ERM trained neural network exhibits high instability over predictions for the OOD data. In contrast, the direct analogue, the EXPLOR neural network, our multi-headed architecture trained using eq.~\ref{eq:head_match}, yields a substantial reduction in variance.
The full EXPLOR model (also bagging with the original pseudolablers) further decreases variance across all datasets, achieving 6–8 times lower variance than ERM. These results provide direct empirical evidence that EXPLOR produces markedly more stable and reliable predictions on unseen chemical scaffolds, and further supports the variance-reduction motivation introduced in  Sec.~\ref{sec:var_reduction}.

\subsubsection{\textbf{Predictive Diversity}}
Beyond model-fitting stability, effective virtual screening also requires structural \emph{diversity} among top-ranked candidates. A useful model should not simply prioritize compounds clustered in a narrow region of chemical space that closely resembles the training data, but instead identify multiple distinct chemotypes that may exhibit the desired activity.
To quantify strutural diversity, we measure the mean variance of ECFP6
fingerprint features among holdout out-of-distribution (OOD) compounds with predicted confidence \(>0.9\) across the three ChEMBL datasets (\(\mathrm{var}@p>0.9\)). EXPLOR achieves the highest value (\textbf{0.39}), compared with D-BAT (0.33), EoA (0.30), AdvStyle (0.34), Mixup (0.37), and NCDG (0.23). Higher fingerprint variance indicates that the model assigns high confidence to compounds with more heterogeneous structural features, suggesting broader exploration of chemical space rather than repeated selection of closely related analogues. Additionally, we examined scaffold diversity by computing the number of unique Bemis–Murcko scaffolds among high-confidence predictions (top 2.5\%). EXPLOR produced the largest number of unique scaffolds (\textbf{26}), exceeding AdvStyle (22), EoA (21), Mixup (21), and D-BAT (19). Scaffolds define the core structural framework of a molecule and largely determine its shape, functional group arrangement, and physicochemical properties. Predictions spanning multiple scaffolds therefore represent distinct chemotypes, which can engage a target through different binding modes.



\begin{figure}[h]
    \centering
    \includegraphics[width=.65\linewidth]{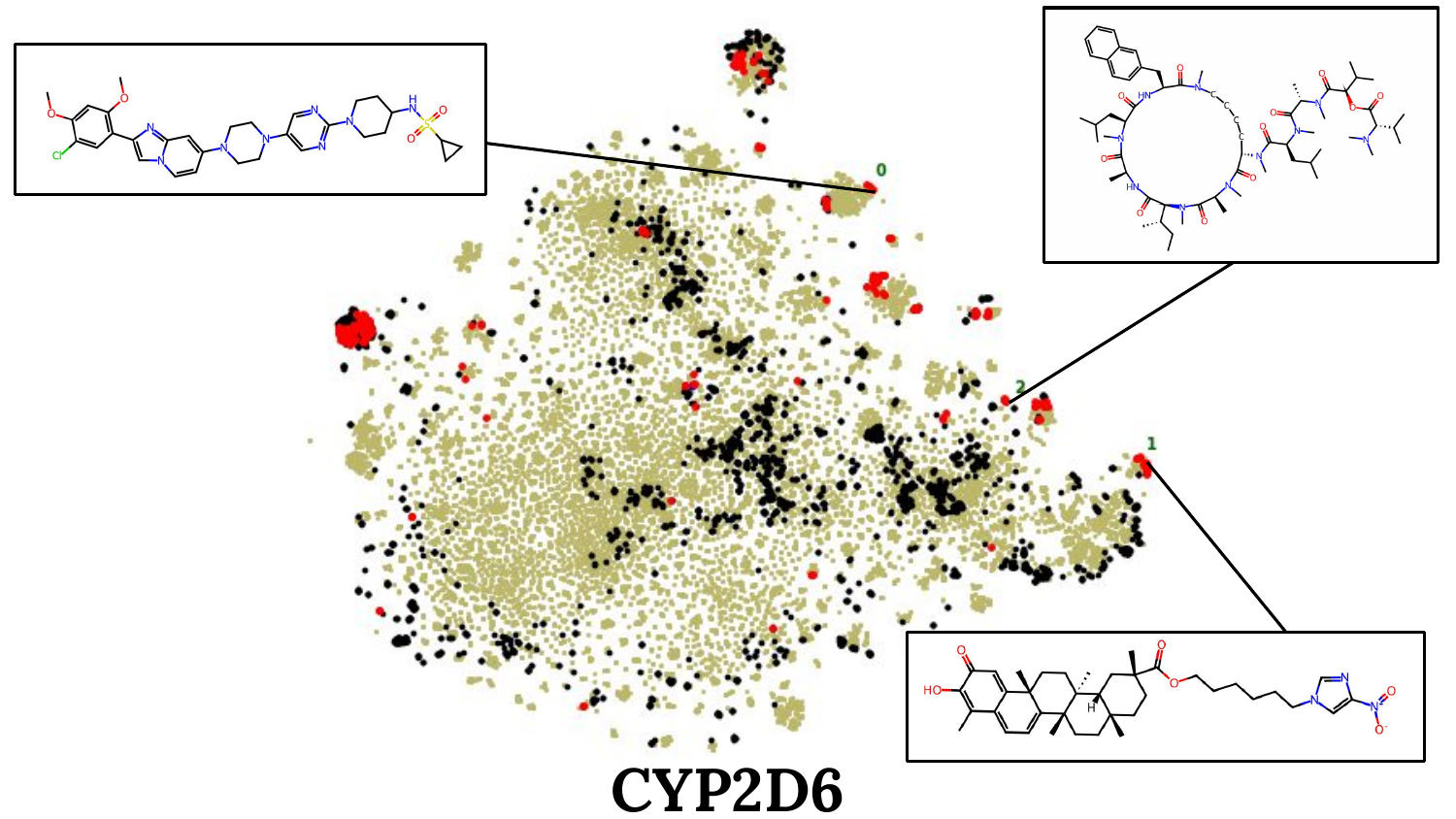}
    \vspace{8pt}
    \includegraphics[width=.65\linewidth]{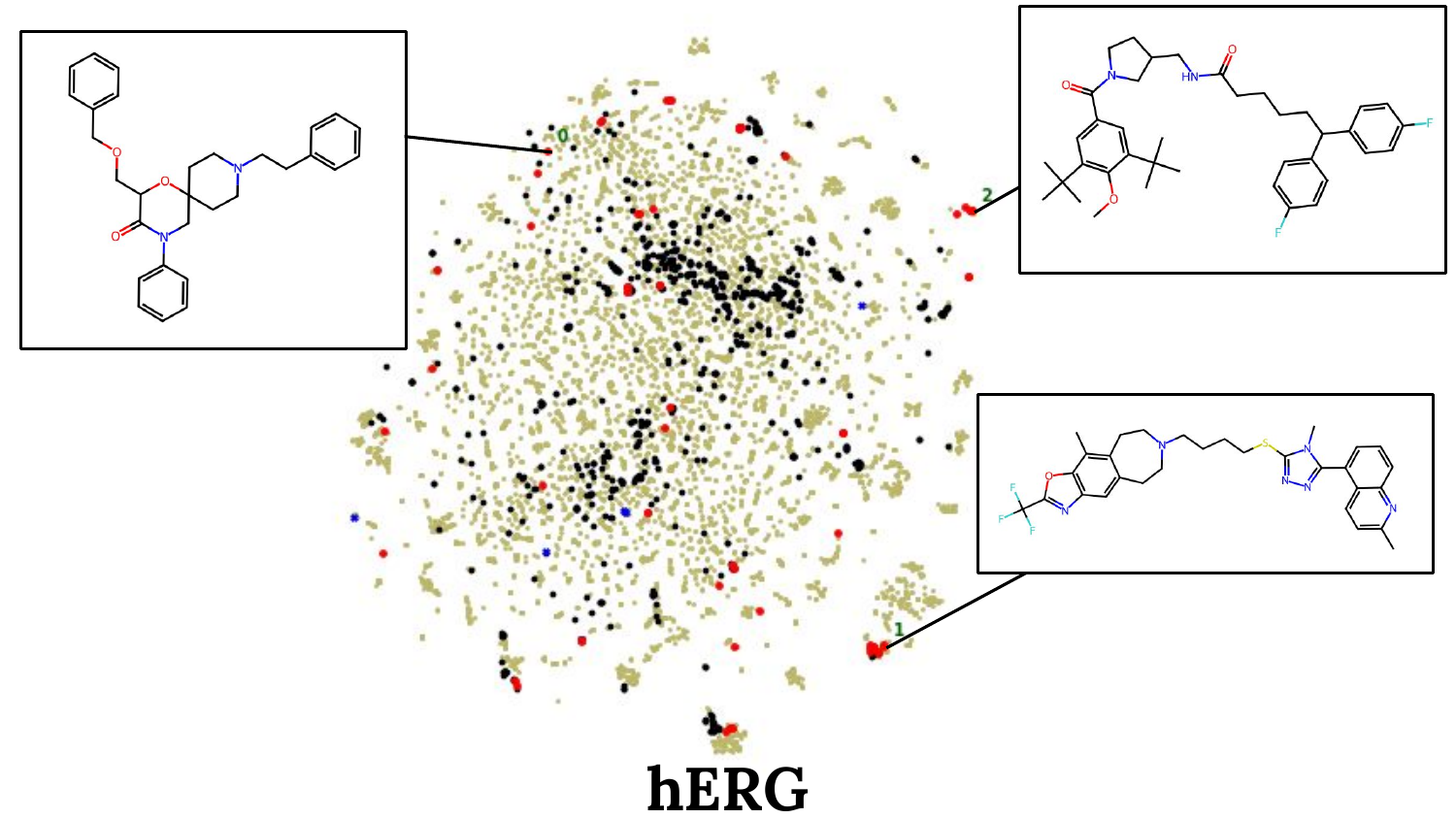}
    \caption{t-SNE visualizations of ECFP6 fingerprints for the CYP2D6 (top) and hERG (bottom) datasets. Training compounds are shown in khaki and candidate compounds in black. The top 10\% of predicted as positive by EXPLOR are highlighted in red (true positives) and blue (false positives). The displayed chemical structures were selected from the top 10 most confident EXPLOR predictions.}
    \label{fig:tsne}
\end{figure}

We examine prediction diversity qualitatively in Fig.~\ref{fig:tsne}, which visualizes EXPLOR’s highest-confidence predictions in a 2D t-SNE representation \cite{JMLR:v9:vandermaaten08a}. High-confidence EXPLOR predictions (red and blue) are distributed across multiple regions of the embedded chemical space rather than collapsing into a single neighborhood of the training distribution (khaki), indicating that the model identifies candidate actives spanning several distinct structural families. 
The representative compounds highlighted in Fig.~\ref{fig:tsne} illustrate the diverse scaffolds and molecule features of EXPLOR candidate molecules. Selected candidates in the hERG dataset include scaffolds with markedly different three-dimensional character, such as a spirocycle with high sp$^3$ character alongside flatter fused-ring systems. Similarly, candidates in the CYP2D6 dataset span substantially different structural classes, including linear heterocycle-containing molecules, a macrocyclic structure, and a terpenoid-like scaffold. These examples demonstrate that EXPLOR prioritizes candidate actives with distinct core architectures and molecular shapes, rather than variations of a single scaffold.


\begin{figure}[h]
    \centering
    \includegraphics[width=0.8\linewidth]{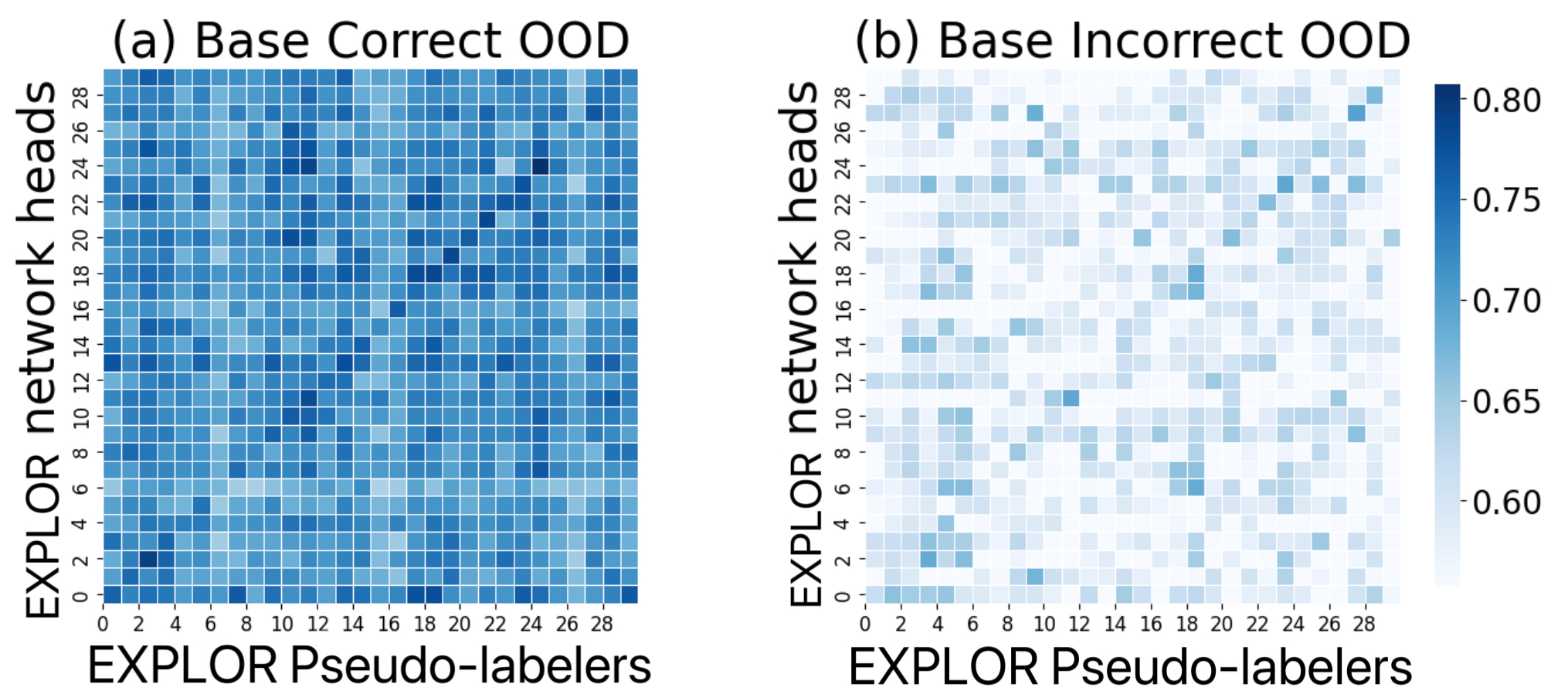}
    \caption{Correlations between EXPLOR pseudo-labelers and EXPLOR network heads on “Ames” dataset. (a) EXPLOR experts have a high correlation with EXPLOR pseudo-labelers on sample the pseudo-labelers make correct predictions. (b) EXPLOR experts show low correlations with EXPLOR pseudo-labelers on samples where the pseudo-labelers make incorrect predictions.}
    \label{fig:expert_corr}
\end{figure}

\subsubsection{\textbf{Ensemble Diversity}}
Finally, we examine diversity within EXPLOR's ensemble of predictors. Ensemble diversity is known to improve performance when members make partially uncorrelated errors \citep{Fort2019DeepEA}, but in EXPLOR the key question is where that diversity appears. Ideally, the model should align strongly with pseudo-labelers when they are reliable, and diversify when they are not. Fig.~\ref{fig:expert_corr} shows this behavior: on OOD samples where pseudo-labelers are correct, EXPLOR heads exhibit high correlation with them, indicating stable recovery of reliable signals; when pseudo-labelers are incorrect, these correlations become weaker and more diffuse, reflecting reduced consensus and greater diversity.

This pattern suggests that EXPLOR does not simply imitate its pseudo-labelers. Instead, its \emph{bottlenecking mechanism and data expansion} aligns with them when they provide useful guidance and decouples when their predictions are likely to be wrong. This error-sensitive diversity helps explain the observed OOD gains, preserving the benefits of extrapolatory supervision while reducing the risk of confidently reproducing pseudo-labeler errors.

\subsection{Ablations and Variants}
\label{sec:ablations}

\subsubsection{\textbf{Varying Molecular Embeddings}}
We next test whether EXPLOR's gains persist across molecular representations. We compare EXPLOR against three baselines using ECFP6 fingerprints, ChemBERTa embeddings \cite{Chithrananda2020ChemBERTaLS}, and MolFormer embeddings \cite{ross2022largescalechemicallanguagerepresentations} on four benchmarks: hERG, A549 Cells, CYP2D6, and Ames.

\begin{figure}[t]
    \centering
    \includegraphics[width=\linewidth]{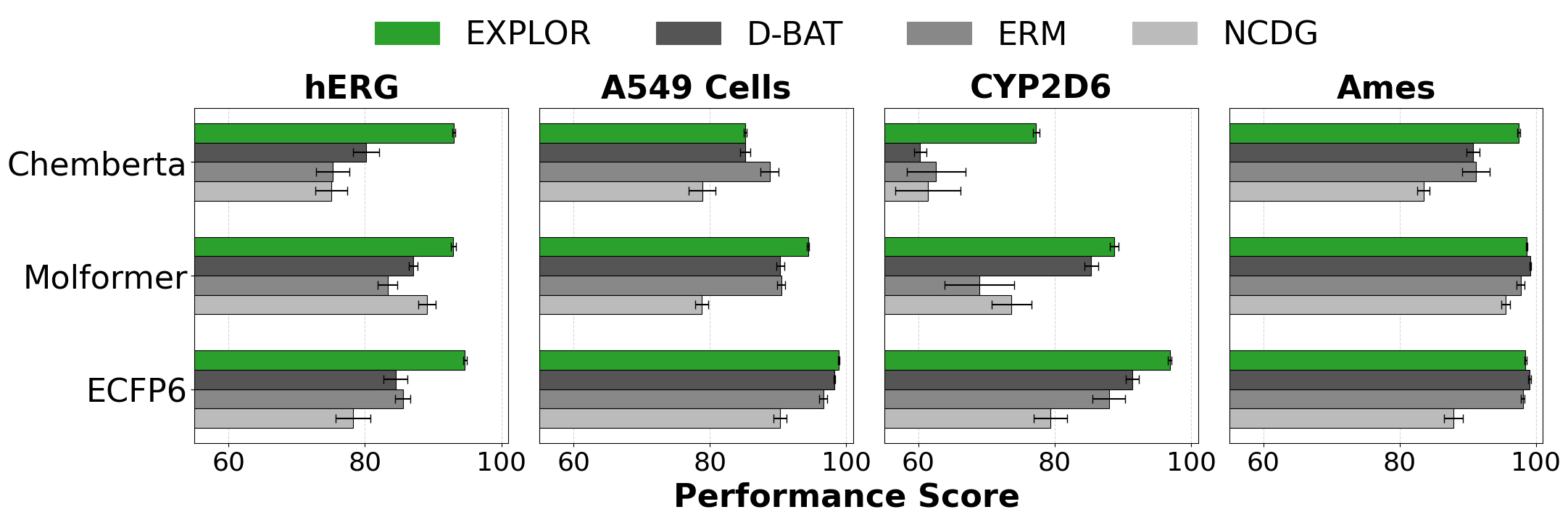}
    
    \caption{Experiment results on Chemberta and Molformer features. Each bar represents the AUPRC@R<0.2 with error bars indicating the std. error across 8 independent runs.}
    \label{fig:horizontal_plots}
\end{figure}

Fig.~\ref{fig:horizontal_plots} shows that EXPLOR is consistently strong across all three feature families and generally attains the best mean $\mathrm{AUPRC}@R{<}0.2$. The gains are especially notable on hERG and CYP2D6, where EXPLOR maintains a clear advantage across ChemBERTa, MolFormer, and ECFP6 features, and remain evident on Ames, where performance is uniformly high but EXPLOR still matches or exceeds the strongest competitors. On A549 Cells, the differences are smaller for some feature types, but EXPLOR remains competitive and is again strongest for MolFormer features. Across settings, EXPLOR also exhibits low run-to-run variability, indicating that its gains are not due to isolated favorable runs.

These results suggest that EXPLOR's advantage is not tied to a single encoder or feature modality. Rather, the method appears to operate effectively on both learned sequence-based molecular embeddings and classical fingerprint representations. This is important for practical screening pipelines, where the preferred representation may vary by dataset, compute budget, or available pretrained models. Overall, the ablation indicates that EXPLOR is relatively robust to the choice of molecular embedding and that its improvements arise from the learning framework itself rather than from dependence on a particular feature extractor.



\subsubsection{\textbf{Cluster-Defined OOD Splits}}

Our main experiments use human-defined OOD splits, such as Bemis--Murcko scaffolds, which capture important but not exhaustive axes of molecular shift. To test whether EXPLOR's gains depend on those handcrafted definitions, we also evaluate a data-driven OOD protocol based on unsupervised clustering.\looseness-1

\emph{Experimental Protocol.} \quad
We run leave-one-out cluster  experiments \cite{kramer2010leave} using K-Means (\(k=5\)) on 128-dimensional PCA representations. Because positives are the minority class and the primary targets in screening, clustering is first performed on positive compounds only. Negative compounds are then assigned to their nearest cluster centroid, yielding five disjoint clusters over the full dataset. Each fold holds out one cluster as OOD test data and trains on the remaining four; results are reported as cluster-size-weighted averages across folds.

\begin{table}
\normalsize
\centering
\caption{Average Performance for Leave-One-Out Cluster Experiments over 3 trials. We \textbf{bold}  best scores based on the mean minus 1 standard error.}
\tiny
\label{tab:loo}
\begin{tabular}{llcccccc}
\toprule
                    &           & AUPRC@  & AUPRC  & AUROC \\
                    &           & {R$<$0.2}   &   &    \\
\midrule
hERG                    & ERM        & 91.88$\pm$0.84 & 75.94$\pm$0.71 & 73.81$\pm$0.68\\
                        & UDIM       & 92.80$\pm$0.76 & 76.56$\pm$0.65 & 73.93$\pm$0.72 \\
                        & D-BAT       & 91.25$\pm$0.91 & 77.79$\pm$0.58 & 76.19$\pm$0.63 \\
                        & AdvStyle   & 89.60$\pm$1.02 & 75.11$\pm$0.83 & 73.16$\pm$0.79 \\
                        & NCDG       & 86.50$\pm$1.15 & 74.63$\pm$0.94 & 73.85$\pm$0.81 \\
                        & EXPLOR     & \textbf{93.98}$\pm$\textbf{0.62} & \textbf{78.98}$\pm$\textbf{0.53} & \textbf{77.00}$\pm$\textbf{0.57} \\
\midrule
A549 Cells             & ERM        & 84.19$\pm$1.23 & 66.84$\pm$0.97 & 66.08$\pm$0.88 \\
                        & UDIM       & 81.41$\pm$1.31 & 66.26$\pm$1.04 & 66.23$\pm$0.95 \\
                        & D-BAT       & 82.54$\pm$1.18 & 67.86$\pm$0.89 & 67.58$\pm$0.82 \\
                        & AdvStyle   & 77.84$\pm$1.44 & 64.20$\pm$1.12 & 64.32$\pm$1.07 \\
                        & NCDG       & 83.73$\pm$1.27 & 67.07$\pm$0.93 & 66.36$\pm$0.91 \\
                        & EXPLOR     & \textbf{91.22}$\pm$\textbf{0.78} & \textbf{71.42}$\pm$\textbf{0.67} & \textbf{69.70}$\pm$\textbf{0.74} \\
\midrule
CYP2D6                  & ERM        & 70.55$\pm$1.37 & 58.99$\pm$1.08 & 59.79$\pm$1.02 \\
                        & UDIM       & 73.68$\pm$1.19 & 59.51$\pm$0.97 & 59.61$\pm$0.94 \\
                        & D-BAT       & 75.04$\pm$1.08 & 61.96$\pm$0.88 & 62.83$\pm$0.85 \\
                        & AdvStyle   & 70.67$\pm$1.42 & 59.13$\pm$1.14 & 60.26$\pm$1.09 \\
                        & NCDG       & 66.22$\pm$1.53 & 57.63$\pm$1.21 & 58.97$\pm$1.17 \\
                        & EXPLOR     & \textbf{79.66}$\pm$\textbf{0.93} & \textbf{63.72}$\pm$\textbf{0.76} & \textbf{64.46}$\pm$\textbf{0.81} \\
\midrule
Ames                    & ERM        & 94.54$\pm$0.61 & 77.39$\pm$0.54 & 72.77$\pm$0.58\\
                        & UDIM       & 88.99$\pm$0.89 & 75.88$\pm$0.72 & 73.14$\pm$0.67 \\
                        & D-BAT       & 90.04$\pm$0.77 & 78.57$\pm$0.61 & 76.24$\pm$0.55 \\
                        & AdvStyle   & 85.77$\pm$1.08 & 74.11$\pm$0.88 & 71.90$\pm$0.84 \\
                        & NCDG       & 83.97$\pm$1.17 & 74.88$\pm$0.91 & 73.31$\pm$0.87 \\
                        & EXPLOR     & \textbf{95.52}$\pm$\textbf{0.48} & \textbf{79.74}$\pm$\textbf{0.43} & \textbf{76.24}$\pm$\textbf{0.52} \\
\bottomrule
\end{tabular}
\normalsize
\end{table}

\emph{Results.}\quad
Table~\ref{tab:loo} shows that EXPLOR is the top-performing method on every dataset and metric under this alternative OOD definition. The gains are most pronounced for \(\mathrm{AUPRC}@R{<}0.2\), where EXPLOR achieves the best score on hERG, A549 Cells, CYP2D6, and Ames, with especially large margins on A549 Cells and CYP2D6. The same pattern extends to full AUPRC and AUROC, where EXPLOR also ranks first throughout.

These findings are important because the held-out regions are not defined by domain heuristics, but emerge directly from the structure of the representation space. Thus, EXPLOR's advantage is not tied to a particular benchmark split construction. Instead, the method continues to improve prioritization of top-ranked screening candidates even when OOD subpopulations are defined automatically. This provides complementary evidence that EXPLOR is learning a more transferable extrapolatory rule, rather than merely adapting well to a specific manually designed notion of shift.

\begin{table}
\centering
\caption{ChEMBL datasets' mean AUPRC at Recall<0.2 varying family of pseudo-labelers.}
\label{tab:base_exp} 
\
\tiny
\begin{tabular}{lcccc}
\toprule
AUPRC@R\textless{}0.2 & XGBoost &D-BAT & Random Forest & Decision Tree \\
\midrule
PL Ens             & 95.87   &91.62& 95.72         & 92.18         \\
EXPLOR                  & \textbf{96.64}  &\textbf{93.81} & \textbf{96.15}         & \textbf{94.15}       \\
\bottomrule
\end{tabular}
\end{table}

\subsubsection{\textbf{Varying Pseudo-Labeler Family}}
\label{sec:varypl}

We next vary the pseudo-labeler family to assess EXPLOR's ability to improve over different base pseudo-labelers. Table~\ref{tab:base_exp} reports mean $\mathrm{AUPRC}@R{<}0.2$ across the ChEMBL datasets using XGBoost, D-BAT, Random Forest (100 estimators), and Decision Tree pseudo-labelers. Across all families, EXPLOR consistently improves over the corresponding pseudo-labeler ensemble baseline. For example, $\mathrm{AUPRC}@R{<}0.2$ increases from 95.87 to 96.64 with XGBoost pseudo-labelers and from 91.62 to 93.81 with D-BAT. This suggests that EXPLOR's gains arise from the training framework itself rather than dependence on any specific pseudo-labeler architecture, with especially large improvements for weaker pseudo-labelers.

\section{Conclusion}
\label{sec:diss}

Machine learning models for virtual screening face a central tension: the compounds most worth discovering often lie in out-of-distribution (OOD) regions beyond the training data, yet predictive performance and confidence estimation typically degrade under such shift. Standard novelty-rejection strategies preserve reliability by abstaining on OOD inputs, but in doing so they can reject precisely the novel scaffolds most valuable for discovery. We introduced EXPLOR, a framework for reliable OOD virtual screening based on extrapolatory pseudo-label matching. By training a multi-headed network to match a diverse set of pseudo-labelers on latent-space augmentations, EXPLOR learns representations that support extrapolation beyond the observed training distribution.

Across scaffold-based ChEMBL benchmarks, DrugOOD ligand-based affinity prediction tasks, and ablations over molecular embeddings, pseudo-labeler families, and alternative OOD split constructions, EXPLOR shows clear gains in the high-confidence early-recall regime most relevant to practical screening, where only a small fraction of top-ranked candidates can be advanced to experimental validation. These gains are accompanied by competitive full AUPRC and AUROC, indicating that EXPLOR improves top-of-list prioritization without sacrificing broader ranking quality. Together, these results position EXPLOR as a simple and flexible approach for improving OOD generalization and reliable candidate prioritization in ligand-based virtual screening under realistic prospective-screening conditions.

\bibliography{bibliography}
\bibliographystyle{ACM-Reference-Format}
\clearpage
\appendix{APPENDIX}
\section{Usage of LLMs Statement}
In this submission, LLMs were used only as a general editing tool. Part of the text were drafted by authors and fed into LLMs for grammar check and help polish the text.

\section{Additional Experiment Details}
\label{sec:appdeats}

\subsection{EXPLOR training details} \label{emoe_hp}
In all of our experiments we used the Adam \citep{Kingma2014AdamAM} optimizer and mini-batches of size 256. One Nvidia A100 GPU with 40GB GPU memory was used to run our experiments, and duration for model training is approximately 0.5 hours.  $\lambda$ = 0.5 was used for the $\mathcal{L}_\mathrm{match}$ for the expanded points. As noted in Sec.~\ref{sec:expaug} we trained the EXPLOR models directly in the latent space to avoid the need for the decoder (and also allowed baselines to do this if it aided their performance). In the experiments on hERG, A549\_cells, CYP\_2D6, Ames, core ec50, refined ec50, EXPLOR was trained for 10000 iterations. Arithmetic mean between EXPLOR and pseudo-labeler ensemble was reported.  We performed 5 trails on each of the datasets for EXPLOR.

\subsection{Baseline Setup}
We implemented all baselines we are comparing against EXPLOR following the implementation details in their paper and/or using Github implementations (if available). Since the fingerprints representation of chemicals are quite sparse, we preformed dimension reduction using PCA with 128 components on all chemical datasets. For D-BAT \citep{Pagliardini2022AgreeTD} with existing implementations designed for tabular data, we utilized their original model architectures. For the other baseline methods without implementation specifically for tabular data, we adopted a structure comprising two 512 ELU\citep{Clevert2015FastAA} layers to closely mimic the EXPLOR network architecture. The Adam \citep{Kingma2014AdamAM} optimizer was used for training baseline models.

\textbf{ERM} We implement a multiheaded ERM baseline that follows the EXPLOR neural network architecture. The architecture uses the same shared feature extractor followed by a 1024-dimensional output layer, where each output corresponds to an independent binary classifier. We train for 10,000 iterations with a learning rate of 0.0005 and maintain a moving average model updated every 2,500 iterations.

\textbf{D-BAT} In our experiments, the D-Bat\citep{Pagliardini2022AgreeTD} models used MLP architecture with one 128 LeakyRelu\citep{Maas2013RectifierNI} layer following the architecture in their Github. Their paper \citep{Pagliardini2022AgreeTD} discussed two settings, and we focused on the scenario where perturbation data differs from the distribution of test data, adhering to the single-source domain generalization setting. We trained an ensemble of five models sequentially for the D-bat baseline models and the predictions from the 5 models were averaged to obtain the final prediction.

\textbf{EoA} We trained an ensemble of 5 simple moving average model following the method described in \citep{Arpit2021EnsembleOA}. We start calculating the moving average  at iteration 50 and trained the models for 200 iterations. The predictions from the 5 models were averaged to obtain the final prediction for EoA.

For \textbf{AdvStyle} \citep{Zhong2022AdversarialSA} and \textbf{Mixup} \citep{Zhang2017mixupBE}, the methodologies were straightforward. We experimented with training using various numbers of iterations and reported the most promising results. Note that we used alpha=0.7 when combining the 2 samples for Mixup. We executed all baseline experiments five times on each dataset to ensure a precise estimation of performance.

\textbf{DivDis}  We utilized all unlabeled target OOD data for training and a single label from this data for supervision. An ensemble of 5 models was trained for 100 iterations with early stopping, and each model has 2 classification heads. Across all datasets, we set $\lambda_1 = 10$ (encouraging disagreement among model heads), while $\lambda_2$ (an optional hyperparameter prevents degenerate solutions) was set to 0 for DrugOOD and ChEMBL and to 10 for TableShift. The final prediction is the average of the 5 models' predictions.

\textbf{FixMatch} Originally, Fixmatch \citep{sohn2020fixmatch} was designed for image data, so the sense of weak and strong augmentations were image based. To adapt the method to our modality agnostic setting we used $x * (1+\alpha)$ as the weak augmentation, $x * (1+2*\alpha)$ as strong augmentation, where $\alpha$ is a small noise drawn from the standard normal distribution.

For \textbf{NCDG} \cite{9730006}, we adapt the method to use the EXPLOR architecture (rather than a ResNet model) and the EXPLOR augmentation method. We set t=0.005 (the threshold for neuron activation in coverage computation), $\lambda = 0.1$ (the weight coefficient for neuron coverage loss), and $\beta = 0.01$ (the weight for gradient similarity regularization loss). Five trials were run on each dataset and averaged to obtain the final results.

\textbf{SAM} We train the MLP with 2 hidden layers of size 512 (same hidden layer as EXPLOR) using the SAM \citep{foret2021sharpnessaware} objective. We used a $\rho=0.05$ (radius for evaluating the loss sharpness) and $\epsilon-1e^{-5}$ (perturbation weight). The model was trained for 100 epochs with a learning rate of 0.001.

\textbf{UDIM} We train the MLP with 2 hidden layers of size 512 (same hidden layer as EXPLOR) using the UDIM \citep{shin2024unknown} framework. We used a $\rho=0.05$ (radius for evaluating the loss sharpness), $\rho_x=0.5$ (radius for adversarial perturbations), and $\lambda = 0.5$ (domain inconsistency regularizer weight). The model was trained for 100 epochs with a learning rate of 0.001.

\subsection{EXPLOR training time} \label{sec:run_time}
In Tab.~\ref{tab:run_time}, we report the training time for EXPLOR and baseline models we are considering. 
Note that the pseudo-labelers can be training in parallel with enough computational resources (and are each quick to train at $<1$s).

\begin{table}[]
\scriptsize
\centering
\caption{Training Time for EXPLOR and baseline methods. }
\begin{tabular}{ll}
\toprule
         & wall clock (m) \\
\midrule
EXPLOR   & 4.2                       \\
SAM      & 1.13\\
UDIM     & 2.07            \\
Dbat     & 3.5                       \\
EoA      & 67.9                      \\
Advstyle & 4.8                       \\
Mixup    & 1.2                       \\
NCDG     & 2.7                      \\
ERM      & 0.9                      \\
\bottomrule
\end{tabular}
\label{tab:run_time}
\end{table}

\section{Additional Experiment and Ablation Results}\label{sec:full_results}

\begin{table*}[htb!]
\centering
\caption{Experiment results on ChEMBL \citep{Gaulton2011ChEMBLAL}, Therapeutics Data Commons \citep{huang2021therapeutics}, and DrugOOD \citep{ji2022drugood} datasets. We \textbf{bold} best scores based on the mean minus 1 standard error. We \textit{italicize} best scores when they are achieved by a semi-supervised method (*), that uses additional unlabeled OOD data during training.}

\tiny
\setlength{\tabcolsep}{3pt}
\begin{tabular}{llllllllllll}
\toprule
                           &           & hERG       & A549\_cells & cyp\_2D6   & Ames  & refined & refined & core & core& core     & core   \\
                           &           &            &             &            &       & ec50 val &ec50 test & ic50 val &ic50 test &ec50 val       &ec50 test  \\
\midrule
\multirow{3}{*}{\rotatebox{90}{AUPRC@R$<$0.2}}
                  & DivDis$^{*}$    & 85.65±2.37 & 89.45±1.16 & 79.98±1.10 & 95.80±0.34 & 96.02±0.19 & 86.51±0.28 & 97.80±0.07 & 89.72±0.06& 93.86±0.42 & 80.18±0.92  \\

                            & FixMatch$^{*}$ & 79.66±0.57 & 90.81±4.30 & 83.58±1.98 & 93.55±0.64 & 94.48±0.47 & 86.38±0.35 & 97.62±0.06 & 92.89±1.36& 95.02±0.76 & 70.49±0.43  \\ 
\cdashline{2-12} \vspace{-6pt} \\
   
                            & ERM             & 85.56±1.11 & 96.65±0.59 & 87.93±2.41 & 98.04±0.30 & 96.15±0.16 & 88.67±0.23 & 99.04±0.04 & 91.44±0.43 & 96.95±0.11 & 72.88±0.95 \\
                            & D-BAT           & 84.48±1.74 & 98.26±0.14 & 91.40±0.98 & 99.04±0.24 & 96.97±0.16 & 88.78±0.40 & 98.13±0.08 & 91.79±0.38 & 93.81±0.22 & \textbf{84.35±1.35} \\
                            & AdvStyle        & 88.21±0.77 & 97.77±0.28 & 84.83±0.93 &99.05±0.17 & 95.13±0.13 & 88.21±0.37 & 97.04±0.17 & 89.05±0.22 & 94.84±0.31 & 84.51±2.36 \\
                            & EoA             & 63.80±0.42 & 61.31±0.31 & 61.77±0.41 & 78.74±0.43 & 85.03±0.06 & 78.79±0.14 & 88.56±0.05 & 77.03±0.13 & 81.85±0.24 & 71.84±0.45 \\
                              & Mixup           & 82.25±1.51 & 95.04±0.25 & 87.09±2.32 & 91.02±1.06 & 85.39±0.23 & 79.78±0.34 & 88.99±0.43 & 78.07±0.61 & 83.97±0.61 & 73.04±0.42 \\
                            & NCDG            & 78.25±2.60 & 90.33±0.95 & 79.31±2.44 & 87.86±1.41 & 93.92±0.62 & 84.46±0.67 & 97.83±0.08 & 87.82±0.26 & 93.40±0.78 & 80.09±1.94 \\
                            & SAM             & 84.37±1.06 & 95.70±0.75 & 85.02±0.69 & 96.66±0.38 & 96.79±0.17 & 87.50±0.23 & 98.78±0.09 & 89.40±0.40 & 95.02±0.37 & 71.92±1.71 \\
                            & UDIM           & 85.07±2.32 & 95.87±0.36 & 84.80±0.84 & 96.53±0.61 & 96.97±0.13 & 88.16±0.78 & 98.61±0.12 & 90.13±0.38 & 94.05±0.30 & 71.37±1.00 \\
                            
\cline{2-12} \vspace{-6pt} \\
                            & D-BAT PL Ens  & 84.86±0.82 & 97.45±0.32 & 92.56±0.81 & 99.47±0.14 & 96.67±0.06 & 88.11±0.04 & 98.74±0.07& 94.05±0.08 & 97.51±0.21 & 69.93±0.04 \\
                           & XGB PL Ens & 94.44±0.17 & 98.22±0.08  & 95.51±0.19 & 97.84±0.20 & 98.00±0.07& 89.48±0.24 & \textbf{99.14±0.02} & 94.20±0.15 & 97.79±0.11        & 68.48±0.30\\ 
                            & \textbf{EXPLOR} (D-BAT)     & 87.05±0.75 & 99.05±0.12 & 95.33±0.51 & \textbf{99.87±0.16} & 97.78±0.04 & 88.97±0.02 & 99.10±0.16 & \textbf{94.52±0.02} & 98.25±0.01 & 75.50±0.15 \\
                           & \textbf{EXPLOR} (XGB)      & \textbf{94.67±0.29} & \textbf{98.87±0.09}  & \textbf{96.88±0.25} & 98.45±0.19 & \textbf{98.45±0.06}& \textbf{89.76±0.26} & 99.15±0.05     & 94.42±0.09 & \textbf{98.66±0.10} & 69.04±0.50  \\

\midrule
\multirow{3}{*}{\rotatebox{90}{AUPRC}}                     
                            & DivDis$^{*}$    & 67.70±0.25 & 76.45±0.63 & 65.76±0.45 & 82.37±0.27 & 89.62±0.12 & 80.92±0.06 & 93.34±0.10 & 81.48±0.30 & 83.85±0.41 & \textit{75.09±0.38} \\
                            & FixMatch$^{*}$  & 45.16±3.11 & 51.31±1.28 & 32.65±2.16 & 72.54±1.15 & 87.14±0.58 & 81.51±0.31 & 92.38±0.15 & 82.78±0.57 & 80.92±1.12 & 70.57±0.32 \\   
\cdashline{2-12} \vspace{-6pt} \\
                            & ERM             & 68.77±0.40 & 81.73±0.37 & 67.97±0.81 & 86.01±0.23  & 89.84±0.07 & 82.26±0.07 & 94.72±0.07 & 82.59±0.21 & 87.70±0.09 & 70.95±0.44 \\
                            & D-BAT           & 54.60±1.59 & 67.04±0.53 & 47.42±0.85 & 70.44±0.76 & 84.70±0.51 & 70.08±0.69 & 90.84±0.28 & 73.45±0.90  & 76.64±0.49 & 54.87±0.99 \\
                            & AdvStyle        & 51.54±0.93 & 65.02±0.72 & 44.41±0.96 & 74.98±0.49 & 83.01±0.90 & 69.48±2.16 & 88.54±1.44 & 72.11±1.49 & 81.17±1.72 & 58.40±2.18 \\
                            & EoA             & 43.30±0.51 & 44.95±0.24 & 37.37±0.95 & 59.43±0.18 & 69.66±0.47 & 57.71±0.79 & 79.12±0.09 & 56.52±0.42  & 64.16±0.51 & 36.50±1.48 \\
                            & Mixup           & 42.42±0.85 & 50.52±0.50 & 27.79±1.49 & 60.94±0.80 & 80.36±0.88 & 72.88±1.67 & 86.88±0.14 & 74.99±0.15& 73.03±1.67 & 60.84±4.31  \\
                            & NCDG            & 65.03±0.75 & 79.50±0.79 & 65.31±1.00 & 76.48±0.30 & 89.27±0.22 & 80.54±0.21 & 94.17±0.08 & 81.19±0.11 & 87.72±0.37 & 74.99±0.74 \\
                            & SAM             & 66.89±0.28 & 79.88±0.34 & 67.37±0.31 & 82.50±0.39 & 89.86±0.09 & 81.93±0.20 & 94.16±0.10 & 81.03±0.23 & 86.66±0.20 & 70.92±0.40 \\
                            & UDIM            & 67.23±0.60 & 79.98±0.37 & 66.86±0.43 & 82.97±1.05 & 89.95±0.11 & 82.19±0.41 & 94.05±0.10 & 81.29±0.25 & 86.35±0.15 & 71.20±0.35 \\


\cline{2-12} \vspace{-6pt} \\
                            & DBAT PL Ens     & 72.22±0.18 & 83.79±0.31 & 73.26±0.65 & 89.08±0.11  & 91.19±0.08 & 82.72±0.04 & 94.79±0.10 & 84.11±0.01 & 88.36±0.28 & 72.13±0.07 \\

                           & XGB PL Ens & 72.19±0.07 & 84.10±0.01  & 72.93±0.13 & 87.43±0.02 & 91.21±0.02  & 82.61±0.05  & 94.91±0.03     & 84.13±0.06 & 88.44±0.05 & 71.80±0.07\\
                            & \textbf{EXPLOR} (D-BAT)     & 72.73±0.33 & \textbf{84.80±0.11} & 73.64±0.45 & \textbf{89.38±0.06}  & 90.56±0.03 & 82.95±0.06 & 94.87±0.13 & 84.11±0.01 & 88.44±0.02 & \textbf{73.13±0.02}  \\
                           & \textbf{EXPLOR} (XGB)     & \textbf{73.26±0.08} & 84.60±0.08  & \textbf{73.59±0.15} & 88.50±0.10 & \textbf{91.59±0.03}& \textbf{83.06±0.10} & \textbf{95.38±0.01}& \textbf{84.77±0.02}  & \textbf{89.52±0.05}& 71.41±0.11     \\
\midrule
\multirow{3}{*}{\rotatebox{90}{AUROC}}                 & DivDis$^{*}$    & 71.19±0.57 & 72.20±0.41 & 64.50±0.77 & 77.05±0.47 & 65.41±0.81  & 58.53±0.46 & 66.68±0.28 & 57.04±0.08 & 73.23±0.40 & 61.15±0.50 \\
                            & FixMatch$^{*}$  & 68.41±0.84 & 61.16±1.31 & \textit{80.85±0.40} & 70.32±0.59 & 70.32±0.59 & 52.69±0.69 & 66.37±0.50 & 58.37±0.58 & 74.38±0.16 & 62.05±0.55 \\   
\cdashline{2-12} \vspace{-6pt} \\
                            & ERM            & 73.58±0.23 & 76.58±0.30 & 65.26±0.55 & 82.02±0.26 & 67.73±0.18 & 59.15±0.23 & 77.41±0.24 & 62.68±0.31 & 72.24±0.09 & 52.32±0.59 \\
                            & D-BAT           & 76.58±0.45 & 78.16±0.23 & 67.54±0.47 & 83.82±0.15 & 75.26±0.28 & 58.21±0.26 & 72.09±0.19 & 60.32±0.25 & \textbf{80.31±0.08} & 64.82±0.18 \\
                            & AdvStyle        & 75.84±0.46 & 76.13±0.28 & 65.51±0.62 & 85.56±0.71 & 75.97±0.39 & 58.86±0.25 & 70.78±0.35 & 59.62±0.30 & 78.36±0.23 & 64.14±0.27 \\
                            & EoA             & 68.02±0.34 & 68.33±0.24 & 60.50±0.48 & 74.77±0.25 & 64.91±0.34 & 52.71±0.44 & 59.27±0.20 & 54.63±0.24 & 62.99±0.16 & 55.83±0.18 \\
                            & Mixup           & 73.96±0.25 & 76.57±0.42 & 67.53±0.90 & 78.43±0.49 & 68.20±0.66 & 56.33±0.45 & 60.39±0.40 & 56.50±0.37 & 64.24±1.23 & 57.75±0.80 \\ 
                            & NCDG            & 70.76±0.20 & 76.45±0.65 & 63.60±0.62 & 73.39±0.68 & 73.66±0.53 & 57.70±0.81 & 67.20±0.33 & 57.18±0.18 & 76.48±0.22 & 61.44±0.13 \\
                            & SAM             & 72.17±0.41 & 74.67±0.49 & 64.45±0.53 & 78.51±0.47 & 67.33±0.22 & 58.97±0.40 & 75.63±0.35 & 60.07±0.35 & 71.06±0.27 & 52.50±0.36 \\
                            & UDIM            & 73.17±0.38 & 74.64±0.38 & 64.09±0.66 & 78.63±1.30 & 67.50±0.29 & 59.30±0.63 & 75.38±0.34 & 60.29±0.39 & 71.37±0.18 & 53.43±0.43 \\

\cline{2-12} \vspace{-6pt} \\
                            & DBAT PL Ens     & 76.51±0.07 & 78.34±0.35 & \textbf{70.80±0.36} & 84.37±0.08  & 69.40±0.06 & 60.04±0.31 & 77.83±0.40 & 64.66±0.04& 74.51±0.48 & 56.39±0.20 \\
                           & XGB PL Ens       & 74.74±0.06 & 79.17±0.02  & 70.33±0.07 & 81.87±0.04 & 73.70±0.07 & 56.48±0.06 &70.17±0.04 &59.77±0.02 & 77.78±0.09  & 64.89±0.06 \\
                            & \textbf{EXPLOR} (D-BAT)     & \textbf{76.84±0.10} & 79.27±0.14 & 70.11±0.33 & \textbf{85.18±0.04}  & 70.64±0.04 & \textbf{59.90±0.03} & \textbf{78.52±0.37} & \textbf{64.67±0.01} & 75.34±0.07 & 56.32±0.03 \\
                           & \textbf{EXPLOR} (XGB)    & 75.97±0.08 & \textbf{79.53±0.06}  & 70.54±0.42 & 83.78±0.12 & \textbf{75.47±0.09}& 55.40±0.20       & 71.28±0.10        & 60.63±0.17 & 80.01±0.03        & \textbf{66.14±0.05}\\
\bottomrule
\end{tabular}

\label{tab:ChEMBL_results}
\end{table*}

\begin{table*}[hbt!]
\tiny
\caption{Full experiment results on DrugOOD datasets. We \textbf{bold} best scores based on the mean minus 1 standard deviation. Note that * refers to a semi-supervised method.}
\label{tab:drugood_results_full}
\centering
\begin{tabular}{llllllll}
\toprule
                     &          & refined ec50 val & refined ec50 test & core ic50 test & core ic 50 test  & core ec50 val       & core ec50 test \\
\midrule
AUPRC@                 & DivDis$^{*}$ & 96.63±0.54 & 88.87±0.58 & 98.41±0.26  & 91.72±1.47 & 98.48±0.06  & 75.02±1.11   \\
{R$<$0.1}            & FixMatch$^{*}$ & 94.78±0.63 & 86.53±0.25 & 97.70±0.05 & 95.89±1.46 & 96.42±1.06 & 69.80±0.25  \\ 
                     & ERM  & 97.76±0.14 & \textbf{90.40±0.27} & 99.38±0.05  & 93.81±0.55& 96.21±1.02  & 82.45±2.93  \\
                     & D-BAT     & 97.19±0.19       & 88.93±0.48        & 98.25±0.09     & 91.89±0.39& 94.04±0.25          & \textbf{86.59±1.42}       \\
                     & AdvStyle  & 96.37±0.32       & 88.69±0.43        & 98.10±0.17     & 89.39±0.33 & 95.79±0.20          & 84.56±2.36          \\
                     & EoA      & 85.03±0.06       & 78.79±0.14        & 88.56±0.05     & 77.03±0.13  & 81.85±0.24          & 71.84±0.45          \\
                     & Mixup     & 85.39±0.23       & 79.78±0.34        & 88.99±0.43     & 78.07±0.61 & 83.97±0.61          & 73.03±0.42           \\
                     & NCDG     & 96.33±0.07       & 89.94±0.12        & 97.36±0.08     & 89.68±0.05   & 91.95±0.28          & 80.38±0.47         \\
                    & SAM             & 98.27±0.15 & 88.84±0.32 & 99.37±0.06 & 91.65±0.41 & 97.29±0.19 & 73.03±2.53 \\
                    & UDIM           & 98.39±0.09 & 89.67±0.95 & 99.19±0.10 & 92.15±0.41 & 96.69±0.25 & 71.10±1.19 \\
\cline{2-8} \vspace{-6pt} \\
                     & D-BAT PL Ens & 98.67±0.02         & 89.48±0.08    & 99.23±0.05       & 96.39±0.07        & 99.01±0.24      & 66.69±0.13     \\
                     & XGB PL Ens & 98.73±0.14          & 65.40±0.52     & \textbf{99.00±0.04}& \textbf{92.67±0.33}& \textbf{99.57±0.01}& \textbf{96.93±0.09}      \\
                     & EXPLOR D-BAT & 98.50±0.01         & 89.94±0.02     & 99.46±0.04      & 96.70±0.01        & 99.93±0.02     & 77.26±0.27     \\
                     & EXPLOR      & \textbf{99.06±0.14}    & 64.71±0.29     & 99.22±0.05       & 91.31±0.41        & 99.36±0.06     & 96.42±0.10    \\
\midrule
AUPRC@               & DivDis$^{*}$  & 96.02±0.19 & 86.51±0.28 & 97.80±0.07 & 89.72±0.06& 93.86±0.42 & 80.18±0.92  \\
{R$<$0.2} 
                     & FixMatch$^{*}$  & 94.48±0.47 & 86.38±0.35 & 97.62±0.06 & 92.89±1.36& 95.02±0.76 & 70.49±0.43  \\ 
                     & ERM             & 96.15±0.16 & 88.67±0.23 & 99.04±0.04 & 91.44±0.43 & 96.95±0.11 & 72.88±0.95 \\
                     & D-BAT           & 96.97±0.16 & 88.78±0.40 & 98.13±0.08 & 91.79±0.38 & 93.81±0.22 & \textbf{84.35±1.35} \\
                     & AdvStyle        & 95.13±0.13 & 88.21±0.37 & 97.04±0.17 & 89.05±0.22 & 94.84±0.31 & 84.51±2.36 \\
                     & EoA             & 85.03±0.06 & 78.79±0.14 & 88.56±0.05 & 77.03±0.13 & 81.85±0.24 & 71.84±0.45 \\
                     & Mixup           & 85.39±0.23 & 79.78±0.34 & 88.99±0.43 & 78.07±0.61 & 83.97±0.61 & 73.04±0.42 \\
                     & NCDG            & 93.92±0.62 & 84.46±0.67 & 97.83±0.08 & 87.82±0.26 & 93.40±0.78 & 80.09±1.94 \\
                     & SAM             & 96.79±0.17 & 87.50±0.23 & 98.78±0.09 & 89.40±0.40 & 95.02±0.37 & 71.92±1.71 \\
                     & UDIM            & 96.97±0.13 & 88.16±0.78 & 98.61±0.12 & 90.13±0.38 & 94.05±0.30 & 71.37±1.00 \\
\cline{2-8} \vspace{-6pt} \\
                     & DBAT PL Ens     & 96.67±0.06 & 88.11±0.04 & 98.74±0.07& 94.05±0.08 & 97.51±0.21 & 69.93±0.04 \\
                     & XGB PL Ens & 98.00±0.07       & 89.48±0.24       & \textbf{99.14±0.02} & 94.20±0.15 & 97.79±0.11        & 68.48±0.30\\ 
                     
                     & EXPLOR DBAT     & 97.78±0.04 & 88.97±0.02 & 99.10±0.16 & 94.52±0.02 & 98.25±0.01 & 75.50±0.15 \\
                     & EXPLOR XGB      & \textbf{98.45±0.06}& \textbf{89.76±0.26} & 99.15±0.05     & \textbf{94.42±0.09} & \textbf{98.66±0.10} & 69.04±0.50  \\
\midrule
AUPRC@               & DivDis$^{*}$  & 95.19±0.37 & 85.12±0.49 & 97.33±0.16 & 88.06±1.14& 91.89±0.95  & 78.97±1.66  \\
{R$<$0.3}            & FixMatch$^{*}$  & 87.14±0.64 & 85.74±0.42 & 97.59±0.07 & 91.02±1.24 & 93.76±1.11 & 70.61±0.47 \\ 
                     & ERM  & 95.06±0.11 & 87.42±0.13 & 98.69±0.05  & 89.86±0.39 & 95.58±0.15  & 71.98±0.84   \\
                     & D-BAT    & 96.89±0.15       & 87.77±0.32        & 98.08±0.08     & 90.71±0.52& 93.73±0.21         & \textbf{81.40±0.89}  \\
                     & AdvStyle  & 94.71±0.13       & 88.05±0.36        & 96.69±0.21     & 88.93±0.19& 94.52±0.37          & 83.57±2.08           \\
                     & EoA       & 85.03±0.06       & 78.79±0.14        & 88.56±0.05     & 77.03±0.13& 81.85±0.24          & 71.84±0.45           \\
                     &Mixup     & 85.39±0.23       & 79.78±0.34        & 88.99±0.43     & 78.07±0.61 & 83.97±0.61         & 73.06±0.43          \\
                     & NCDG       & 95.17±0.04       & 86.07±0.08        & 95.89±0.05     & 86.53±0.03& 88.73±0.19         & 76.05±0.35          \\
                    & SAM            & 95.62±0.13 & 86.48±0.27 & 98.23±0.09 & 87.84±0.37 & 93.58±0.39 & 71.63±1.27 \\
                    & UDIM           & 95.72±0.15 & 87.08±0.77 & 98.10±0.11 & 88.48±0.40 & 92.46±0.35 & 71.53±0.86 \\
\cline{2-8} \vspace{-6pt} \\
                     & D-BAT PL Ens & 96.10±0.06     & 87.35±0.04    & 98.36±0.10            & 92.44±0.05        & 95.86±0.25     & 71.17±0.03     \\
                     & XGB PL Ens & 96.73±0.10        & 69.94±0.20     & 97.17±0.06       & 87.99±0.18        & 98.71±0.03     & 92.32±0.23      \\
                     & EXPLOR D-BAT  & 97.01±0.01     & \textbf{87.87±0.02}     & 98.61±0.07          & 92.78±0.02        & 96.27±0.01     & 74.76±0.10     \\
                     & EXPLOR XGB     & \textbf{97.91±0.07}& 69.74±0.21     & \textbf{97.76±0.05}& \textbf{88.66±0.21} & \textbf{98.89±0.02}& \textbf{93.11±0.13}   \\
\midrule
AUPRC                & DivDis$^{*}$   & 89.62±0.12 & 80.92±0.06 & 93.34±0.10 & 81.48±0.30 & 83.85±0.41 & \textit{75.09±0.38} \\
                     & FixMatch$^{*}$ & 87.14±0.58 & 81.51±0.31 & 92.38±0.15 & 82.78±0.57 & 80.92±1.12 & 70.57±0.32 \\   
                     & ERM            & 89.84±0.07 & 82.26±0.07 & 94.72±0.07 & 82.59±0.21 & 87.70±0.09 & 70.95±0.44 \\
                     & D-BAT         & 84.70±0.51 & 70.08±0.69 & 90.84±0.28 & 73.45±0.90  & 76.64±0.49 & 54.87±0.99 \\
                     & AdvStyle       & 83.01±0.90 & 69.48±2.16 & 88.54±1.44 & 72.11±1.49 & 81.17±1.72 & 58.40±2.18 \\
                     & EoA           & 69.66±0.47 & 57.71±0.79 & 79.12±0.09 & 56.52±0.42  & 64.16±0.51 & 36.50±1.48 \\
                     & Mixup          & 80.36±0.88 & 72.88±1.67 & 86.88±0.14 & 74.99±0.15& 73.03±1.67 & 60.84±4.31  \\
                     & NCDG           & 89.27±0.22 & 80.54±0.21 & 94.17±0.08 & 81.19±0.11 & 87.72±0.37 & 74.99±0.74 \\
                     & SAM            & 89.86±0.09 & 81.93±0.20 & 94.16±0.10 & 81.03±0.23 & 86.66±0.20 & 70.92±0.40 \\
                     & UDIM            & 89.95±0.11 & 82.19±0.41 & 94.05±0.10 & 81.29±0.25 & 86.35±0.15 & 71.20±0.35 \\
\cline{2-8} \vspace{-6pt} \\
                     & DBAT PL Ens     & 91.19±0.08 & 82.72±0.04 & 94.79±0.10 & 84.11±0.01 & 88.36±0.28 & 72.13±0.07 \\
                     & XGB PL Ens & 91.21±0.02       & 82.61±0.05        & 94.91±0.03     & 84.13±0.06      & 88.44±0.05        & 71.80±0.07\\
                     
                     & EXPLOR D-BAT    & 90.56±0.03 & 82.95±0.06 & 94.87±0.13 & 84.11±0.01 & 88.44±0.02 & \textbf{73.13±0.02}  \\
                     & EXPLOR XGB     & \textbf{91.59±0.03}& \textbf{83.06±0.10} & \textbf{95.38±0.01}& \textbf{84.77±0.02}  & \textbf{89.52±0.05}& 71.41±0.11     \\
\midrule
AUROC               & DivDis$^{*}$  & 65.41±0.81  & 58.53±0.46 & 66.68±0.28 & 57.04±0.08 & 73.23±0.40 & 61.15±0.50 \\
                     & FixMatch$^{*}$ & 70.32±0.59 & 52.69±0.69 & 66.37±0.50 & 58.37±0.58 & 74.38±0.16 & 62.05±0.55 \\   
                     & ERM            & 67.73±0.18 & 59.15±0.23 & 77.41±0.24 & 62.68±0.31 & 72.24±0.09 & 52.32±0.59 \\
                     & D-BAT         & 75.26±0.28 & 58.21±0.26 & 72.09±0.19 & 60.32±0.25 & \textbf{80.31±0.08} & 64.82±0.18 \\
                     & AdvStyle      & 75.97±0.39 & 58.86±0.25 & 70.78±0.35 & 59.62±0.30 & 78.36±0.23 & 64.14±0.27 \\
                     & EoA           & 64.91±0.34 & 52.71±0.44 & 59.27±0.20 & 54.63±0.24 & 62.99±0.16 & 55.83±0.18 \\
                     & Mixup         & 68.20±0.66 & 56.33±0.45 & 60.39±0.40 & 56.50±0.37 & 64.24±1.23 & 57.75±0.80 \\ 
                     & NCDG          & 73.66±0.53 & 57.70±0.81 & 67.20±0.33 & 57.18±0.18 & 76.48±0.22 & 61.44±0.13 \\
                     & SAM            & 67.33±0.22 & 58.97±0.40 & 75.63±0.35 & 60.07±0.35 & 71.06±0.27 & 52.50±0.36 \\
                     & UDIM            & 67.50±0.29 & 59.30±0.63 & 75.38±0.34 & 60.29±0.39 & 71.37±0.18 & 53.43±0.43 \\
\cline{2-8} \vspace{-6pt} \\
                     & DBAT PL Ens     & 69.40±0.06 & 60.04±0.31 & 77.83±0.40 & 64.66±0.04& 74.51±0.48 & 56.39±0.20 \\
                     & XGB PL Ens & 73.70±0.07         & 56.48±0.06        &70.17±0.04          &59.77±0.02         & 77.78±0.09  & 64.89±0.06          \\
                     
                     & EXPLOR DBAT     &70.64±0.04 & \textbf{59.90±0.03} &\textbf{78.52±0.37} & \textbf{64.67±0.01} & 75.34±0.07 & 56.32±0.03 \\

                     & EXPLOR XGB     & \textbf{75.47±0.09}& 55.40±0.20       & 71.28±0.10        & 60.63±0.17 & 80.01±0.03        & \textbf{66.14±0.05}          \\
\bottomrule
\end{tabular}
\end{table*}

\subsection{Full Experiment Results on CheMBL and Therapeutics Data Commons}\label{sec:chem_tdc_res}
In Tab.~\ref{tab:ChEMBL_TDC_results_full}, we report the full results on hERG, A549\_cells, cyp\_2D6. and Ames.

\begin{table}[hbt!]
\tiny
\centering
\caption{Full experiment results on ChEMBL \citep{Gaulton2011ChEMBLAL} and Therapeutics Data Commons \citep{huang2021therapeutics} datasets. We \textbf{bold} best scores based on the mean minus 1 standard deviation. Note that * refers to a semi-supervised method.}
\label{tab:ChEMBL_TDC_results_full}

\begin{tabular}{llllll}
\toprule
                           &           & hERG       & A549\_cells & cyp\_2D6   & Ames \\
\midrule
AUPRC& DivDis$^{*}$    &  86.92±8.44 &  89.01±3.53 & 83.65±6.17 & 97.47±1.68\\
{R$<$0.1}                  & FixMatch$^{*}$ & 84.77±1.81 & 92.93±3.00 & 92.71±1.98 & 96.26±1.04  \\  
\cline{2-6} \\
                           & ERM & 91.95±0.85 & 97.37±0.81 & 92.89±2.93 & 99.02±0.15  \\
                           & D-BAT & 88.55±1.68 & 98.57±0.16  & 95.71±0.89 & 99.07±0.23 \\
                           & AdvStyle  & 93.27±0.56 & 96.89±0.30  & 84.21±2.07 & 99.52±0.12 \\
                           & EoA       & 63.80±0.42 & 61.31±0.31  & 61.77±0.41 & 78.74±0.43 \\
                           & Mixup     & 82.80±1.56 & 95.04±0.25  & 87.39±3.09 & 91.02±1.05 \\
                           & NCDG      & 72.96±1.31 & 78.79±2.35  & 61.77±1.15 & 89.68±0.38 \\
                            & SAM             & 90.69±1.67 & 97.22±0.63 & 90.20±0.46 & 98.29±0.19 \\
                            & UDIM           & 91.03±2.36 & 97.39±0.42 & 88.30±1.14 & 97.99±0.44 \\
\cline{2-6} \vspace{-6pt} \\
                            & D-BAT PL Ens & 86.29±1.00         & 98.38±0.23     & 96.66±0.63       & 1.00±0.00        \\
                           & XGB PL Ens & 96.65±0.22 & 99.74±0.05  & \textbf{99.81±0.11} & 98.73±0.27 \\
                           & EXPLOR D-BAT   & 90.01±0.86         & 98.92±0.12     & 98.51±0.63       & 1.00±0.00        \\
                           & EXPLOR XGB      & \textbf{98.10±0.34} & \textbf{99.76±0.03}  &99.48±0.17 & \textbf{99.66±0.20} \\
\midrule
AUPRC@                     & DivDis$^{*}$    & 85.65±2.37 & 89.45±1.16 & 79.98±1.10 & 95.80±0.34\\
{R$<$0.2}                  & FixMatch$^{*}$ & 79.66±0.57 & 90.81±4.30 & 83.58±2.55 & 93.55±0.64  \\
\cline{2-6} \vspace{-2pt} \\
                           & ERM & 85.58±1.10 & 96.65±0.59 & 87.93±2.41 & 98.04±0.30  \\
                           & D-BAT     & 84.48±1.74 & 98.26±0.14  & 91.40±0.98 & 99.04±0.24 \\
                           & AdvStyle  & 88.21±0.77 & 97.77±0.28  & 84.83±0.93 & \textbf{99.05±0.17} \\
                           & EoA       & 63.80±0.42 & 61.31±0.31  & 61.77±0.41 & 78.74±0.43 \\
                           & Mixup     & 82.25±1.51 & 95.04±0.25  & 87.09±2.32 & 91.02±1.05 \\
                           & NCDG      & 70.25±1.03 & 78.22±2.13  & 61.48±0.98 & 86.29±0.45 \\
                            & SAM             & 84.37±1.06 & 95.70±0.75 & 85.02±0.69 & 96.66±0.38 \\
                            & UDIM           & 85.07±2.32 & 95.87±0.36 & 84.80±0.84 & 96.53±0.61 \\
\cline{2-6} \vspace{-6pt} \\
                            & DBAT PL Ens  & 84.86±0.82 & 97.45±0.32 & 92.56±0.81 & 99.47±0.14 \\
                           & XGB PL Ens & 94.44±0.17 & 98.22±0.08  & 95.51±0.19 & 97.84±0.20 \\
                            & EXPLOR DBAT     & 87.05±0.75 & 99.05±0.12 & 95.33±0.51 & \textbf{99.87±0.16} \\
                           & EXPLOR XGB      & \textbf{94.67±0.29} & \textbf{98.87±0.09}  & \textbf{96.88±0.25} & 98.45±0.19 \\
\midrule
AUPRC@                     & DivDis$^{*}$ &83.69±3.39 & 88.83±2.50 & 77.55±1.02 & 94.87±0.74 \\
{R$<$0.3}                  & FixMatch$^{*}$ & 76.16±1.04 & 89.12±5.07 & 77.70.58±2.05 & 92.72±0.51 \\ 
\cline{2-6} \vspace{-2pt} \\
                           & ERM & 82.29±1.10 & 95.92±0.58 & 83.76±2.02 & 97.23±0.36  \\
                           & D-BAT     & 82.44±1.59 & 97.37±0.24  & 87.65±0.58 & 98.61±0.24 \\
                           & AdvStyle  & 85.05±0.86 & 96.47±0.29  & 82.76±0.95 & \textbf{98.71±0.20} \\
                           & EoA       & 63.80±0.42 & 61.31±0.31  & 61.77±0.41 & 78.74±0.43 \\
                           & Mixup     & 81.51±1.35 & 94.95±0.25  & 84.53±2.89 & 90.88±1.00 \\
                           & NCDG      & 69.59±0.97 & 77.88±1.93  & 60.93±0.86 & 85.41±0.33 \\
                            & SAM         & 80.55±0.72 & 94.18±0.61 & 82.66±0.68 & 95.24±0.36 \\
                            & UDIM           & 80.87±1.85 & 94.67±0.46 & 82.03±0.38 & 95.36±0.73 \\
\cline{2-6} \vspace{-6pt} \\
                            & D-BAT PL Ens & 84.13±0.59         & 96.97±0.14     & 89.06±1.24       & 98.77±0.10        \\
                           & XGB PL Ens & 90.46±0.06 & 97.35±0.05  & 92.34±0.24 & 97.83±0.15 \\
                           & EXPLOR D-BAT   & 85.66±0.72         & 98.70±0.02     & 91.40±0.44       & 99.24±0.06        \\
                           & EXPLOR XGB     & \textbf{90.88±0.34} & \textbf{97.96±0.1}  & \textbf{93.06±0.27} & 98.18±0.15 \\
\midrule
{AUPRC}                     & DivDis$^{*}$    & 67.70±0.25 & 76.45±0.63 & 65.76±0.45 & 82.37±0.27 \\
                            & FixMatch$^{*}$  & 45.16±3.11 & 51.31±1.28 & 32.65±2.16 & 72.54±1.15 \\
\cline{2-6} \vspace{-2pt} \\
                           & ERM & 68.77±0.40 & 81.73±0.37 & 67.97±0.81 & 86.01±0.23  \\
                           & D-BAT     & 54.60±1.59 & 67.04±0.53  & 47.42±0.85 & 70.44±0.76 \\
                           & AdvStyle  & 51.54±0.93 & 65.02±0.72  & 44.41±0.96 & 74.98±0.49 \\
                           & EoA       & 43.30±0.51 & 44.95±0.24  & 37.37±0.95 & 59.43±0.18 \\
                           & Mixup     & 42.42±0.85 & 50.52±0.50  & 27.79±1.49 & 60.94±0.80 \\
                           & NCDG      & 42.69±0.24 & 49.13±0.79  & 31.08±0.16 & 67.79±0.93 \\
                            & SAM             & 66.89±0.28 & 79.88±0.34 & 67.37±0.31 & 82.50±0.39 \\
                            & UDIM            & 67.23±0.60 & 79.98±0.37 & 66.86±0.43 & 82.97±1.05 \\
\cline{2-6} \vspace{-6pt} \\
                            & DBAT PL Ens     & 72.22±0.18 & 83.79±0.31 & 73.26±0.65 & 89.08±0.11  \\

                           & XGB PL Ens & 72.19±0.07 & 84.10±0.01  & 72.93±0.13 & 87.43±0.02 \\
                            & EXPLOR DBAT     & 72.73±0.33 & 84.80±0.11 & 73.64±0.45 & \textbf{89.38±0.06}  \\
                           & EXPLOR XGB     & \textbf{73.26±0.08} & \textbf{84.60±0.08}  & \textbf{73.59±0.15} & 88.50±0.10 \\
\midrule
{AUROC}                     & DivDis$^{*}$    & 71.19±0.57 & 72.20±0.41 & 64.50±0.77 & 77.05±0.47 \\
                            & FixMatch$^{*}$  & 68.41±0.84 & 61.16±1.31 & \textit{80.85±0.40} & 70.32±0.59 \\       
\cline{2-6} \vspace{-2pt} \\
                           & ERM & 73.58±0.23 & 76.58±0.30 & 65.25±0.55 & 82.02±0.26  \\
                           & D-BAT     & 76.58±0.45 & 78.16±0.23  & 67.54±0.47 & 83.82±0.15 \\
                           & AdvStyle  & 75.84±0.46 & 76.13±0.28  & 65.51±0.62 & \textbf{85.56±0.71} \\
                           & EoA       & 68.02±0.34 & 68.33±0.24  & 60.50±0.48 & 74.77±0.25 \\
                           & Mixup     & 73.96±0.26 & 76.57±0.42  & 67.53±0.90 & 78.43±0.49 \\
                           & NCDG      & 70.42±0.13 & 72.50±0.79  & 62.01±0.16 & 73.42±0.93 \\
                            & SAM             & 72.17±0.41 & 74.67±0.49 & 64.45±0.53 & 78.51±0.47 \\
                            & UDIM            & 73.17±0.38 & 74.64±0.38 & 64.09±0.66 & 78.63±1.30 \\
\cline{2-6} \vspace{-6pt} \\
                            & DBAT PL Ens     & 76.51±0.07 & 78.34±0.35 & \textbf{70.80±0.36} & 84.37±0.08  \\
                           & XGB PL Ens & 74.74±0.06 & 79.17±0.02  & 70.33±0.07 & 81.87±0.04 \\
                            & EXPLOR DBAT     & \textbf{76.84±0.10} & 79.27±0.14 & 70.11±0.33 & \textbf{85.18±0.04}  \\
                           & EXPLOR XGB     & 75.97±0.08 & \textbf{79.53±0.06}  & 70.54±0.42 & 83.78±0.12 \\
\bottomrule
\end{tabular}
\normalsize

\end{table}

\subsection{Full Experiment Results on DrugOOD}\label{sec:drugood_res}
In Tab.~\ref{tab:drugood_results_full}, we report the full results on core ec50, refined ec 50, and core ic50 from DrugOOD \citep{ji2022drugood}.

\subsection{Other Real World Datasets}


We further evaluate our method in non-chemical domains across a diverse range of real-world OOD scenarios using the Tableshift datasets \citep{gardner2023tableshift}.
We selected a diverse collection of  Tableshift datasets, based on unrestricted availability and in/out-of-domain performance discrepancy, coverings areas including: finance, education, and healthcare. Each dataset has an associated real-world shift and a related prediction target (see \citet{gardner2023tableshift} for further details). 
Results on the Tableshift are shown in Tab.~\ref{tab:tableshift_PACS_results}. 
As before, we consider the same single-source domain generalization setting.
We can see that even over diverse applications, our EXPLOR method is able to perform well and often outperforms competitive baselines. Moreover, eventhough DivDis and FixMatch use additional unlabeled OOD data, EXPLOR outperforms both on the majority of datasets, highlighting EXPLOR’s 
ability for generalization on completely \emph{unforeseen} OOD instances (without any knowledge of test time distribution). 

\begin{table}
\tiny
\caption{Full experiment results on Tableshift \citep{gardner2023tableshift} datasets. We \textbf{bold}  best scores based on the mean minus 1 standard error. Note that * refers to a semi-supervised method that uses additional unlabeled OOD data during training.}
\tiny
\centering
\label{tab:tableshift_PACS_results}
\begin{tabular}{llcccccc}
\toprule
                    &           & Childhood   & FICO  & Hospital   & Sepsis  \\
                    &           & Lead   &HELOC   &Readmission   &   \\

\midrule
AUPRC@                   & DivDis$^{*}$   & 91.67±3.03 & 84.88±3.80 & 88.99±3.36 & 59.50±1.55\\
{R$<$0.1}                  & FixMatch$^{*}$ & 86.68±3.32& 83.99±2.27&73.97±2.64 &17.84±1.87\\
\cline{2-6}\\
                           & ERM        &54.24±0.00 & 85.84±0.88 & \textbf{90.52±0.36} & 17.35±0.94 \\
                           & D-BAT        &52.67±0.00 &92.97±0.59 &83.73±0.12 &81.99±0.33 \\
                           & AdvStyle      &63.63±0.00 &90.10±1.03 &77.29±0.74 &60.79±0.72 \\
                           & EoA           &75.73±0.83 &63.42±2.13 &52.03±1.78 &43.76±1.14 \\
                           & Mixup         & 50.00±0.00&92.79±0.03 &85.71±0.35 &68.21±1.28 \\
                           & NCDG          & 83.30±0.20 & 91.70±0.33 & 70.42±0.17 & 72.40±0.04 \\
                            & SAM             & 97.50±0.00 & 94.42±1.07 & 80.13±0.57 & 66.42±1.20 \\
                            & UDIM           & 97.06±0.45 & 95.07±0.92 & 79.59±1.32 & 65.62±0.53 \\
\cline{2-6} \vspace{-6pt} \\
                            & D-BAT PL Ens & 98.72±0.42         & 96±0.05     & 76.47±0.05       & 75.32±0.01        \\
                           & XGB PL Ens     &98.69±0.03 &92.18±0.58 &51.31±0.16 & \textbf{82.85±1.53} \\
                           & EXPLOR D-BAT   & 98.36±0.05         & 96.99±0.06     & 81.84±0.34       & 77.25±0.17        \\
                           & EXPLOR XGB          &\textbf{99.72±0.10} & \textbf{93.93±0.87} &66.65±0.17 &80.74±1.77 \\
\midrule
AUPRC@                   & DivDis$^{*}$   & 89.77±2.51 & 85.50±2.65 & 83.22±2.57 & 60.66±1.57 \\
{R$<$0.2} & FixMatch$^{*}$ & 81.75±3.69 & 77.76±1.94 & 67.84±1.95 & 17.28±0.58 \\
\cline{2-6}\\
                           & ERM      &43.66±0.00 & 85.74±0.78 & \textbf{84.35±0.28} & 15.31±0.53 \\
                           & D-BAT         & 62.82±0.00 & 91.20±0.24 & 78.84±0.12 & 75.37±0.38 \\
                           & AdvStyle      & 64.96±0.01 & 88.71±0.65 & 72.91±0.58 & 59.83±0.63 \\
                           & EoA           & 77.43±0.97 & 59.53±2.24 & 51.83±1.66 & 41.10±1.56 \\
                           & Mixup         & 50.00±0.00 & 91.16±2.10 & 69.19±3.95 & 66.09±1.15 \\
                           & NCDG          & 82.49±0.18 & 90.93±0.31 & 68.88±0.15 & 70.22±0.03 \\
                           & SAM           & 95.00±0.00 & 91.42±0.69 & 74.13±0.49 & 65.34±0.73 \\
                           & UDIM           & 94.11±0.90 & 92.58±0.67 & 73.65±0.67 & 65.09±0.28 \\
\cline{2-6} \vspace{-6pt} \\
                         & DBAT PL Ens      & 93.81±0.03 & 92.87±0.05 & 73.52±0.37 & 72.74±0.13 \\
                           & XGB PL Ens     & 97.39±0.06 & 90.07±0.32 & 58.30±0.09 & \textbf{78.62±1.57} \\
                         & \textbf{EXPLOR} (D-BAT)      & 94.69±0.02 & \textbf{93.33±0.02} & 77.20±0.23 & 73.89±0.08 \\
                           & \textbf{EXPLOR} (XGB)         & \textbf{97.92±0.20} & 91.72±0.62 & 67.57±0.12 & 76.95±1.45 \\
\midrule
AUPRC@                   & DivDis$^{*}$   & 87.90±2.20 & 85.77±2.73 &  79.82±2.42 & 61.55±1.55 \\
{R$<$0.3}                & FixMatch$^{*}$ &81.61±3.09 &76.48±4.13 &64.47±1.70 &18.45±0.46 \\
\cline{2-6}\\
                           & ERM       &38.83±0.00 &85.25±0.60 & \textbf{80.54±0.25} & 14.41±0.35 \\
                           & D-BAT         &68.11±0.00 &90.11±0.13 &75.26±0.10 &70.92±0.44 \\
                           & AdvStyle      &68.67±0.08 &87.33±1.21 &70.27±0.49 &59.02±0.57 \\
                           & EoA           &79.88±1.03 &56.52±2.34 &51.34±1.27 &40.09±1.05 \\
                           & Mixup        & 50.00±0.00&90.01±0.02 &47.18±3.54 &64.41±0.95 \\
                           & NCDG          & 82.10±0.15 & 88.76±0.29 & 67.52±0.13 & 69.48±0.03 \\
                            & SAM             & 92.50±0.00 & 89.90±0.61 & 71.03±0.44 & 64.28±0.57 \\
                            & UDIM           & 91.17±1.36 & 91.06±0.44 & 70.66±0.60 & 64.24±0.23 \\
\cline{2-6} \vspace{-6pt} \\
                            & D-BAT PL Ens & 91.71±0.04         & 91.52±0.01     & 71.78±0.27       & 70.33±0.16        \\
                           & XGB PL Ens     &96.08±0.08 &89.73±0.23 &60.86±0.04 & \textbf{75.84±1.55} \\
                           & EXPLOR D-BAT   & 92.60±0.03         & 91.72±0.07     & 72.69±0.18       & 70.92±0.06        \\
                           & EXPLOR XGB     &\textbf{96.58±0.22} & \textbf{91.10±0.41} &67.36±0.06 &74.02±1.20 \\
\midrule
AUPRC                    & DivDis$^{*}$   & 76.33±0.86 & 82.47±1.05 & 65.95±2.15 & 58.85±1.21 \\
                         & FixMatch$^{*}$ & 75.39±0.58 & 72.22±4.46 & 56.54±1.21 & 18.24±0.66 \\
\cline{2-6}\\
                         & ERM       &23.60±0.11 & 77.37±0.27 & \textbf{67.56±0.13} & 11.12±0.12 \\
                         & D-BAT          & 71.85±0.01 & 80.91±0.17 & 63.29±0.08 & 58.17±0.21 \\
                         & AdvStyle       & 48.37±2.77 & 79.63±1.65 & 38.13±3.80 & 54.34±0.27 \\
                         & EoA            & 49.48±0.19 & 59.53±2.24 & 29.45±4.95 & 11.21±2.21 \\
                         & Mixup          & 50.00±0.00 & 80.95±0.63 & 14.15±1.06 & 56.80±0.40 \\
                         & NCDG           & 73.08±0.14 & 79.05±0.21 & 58.95±0.10 & 61.96±0.02 \\
                           & SAM           & 75.00±0.00 & 81.61±0.43 & 61.12±0.28 & 57.90±0.22 \\
                           & UDIM           & 73.70±1.38 & 82.10±0.44 & 60.95±0.33 & 57.97±0.17 \\
\cline{2-6} \vspace{-6pt} \\
                         & DBAT PL Ens      & 82.72±0.01 & 81.35±0.03 & 62.34±0.14 & 59.97±0.10 \\
                         & XGB PL Ens        & 86.39±0.19 & 83.80±0.07 & 62.83±0.03 & \textbf{64.01±0.94} \\
                         & \textbf{EXPLOR} (D-BAT)      & 82.76±0.02 & 81.37±0.04 & 63.30±0.14 & 50.97±0.05 \\
                         & \textbf{EXPLOR} (XGB)           & \textbf{86.70±0.28} & \textbf{84.02±0.09} & 63.62±0.08 & 62.21±0.52 \\
\midrule
AUROC                    & DivDis$^{*}$   & 77.74±1.90 & 83.55±1.06 & 65.28±0.59 & 62.33±0.61 \\
                         & FixMatch$^{*}$ & 79.87±0.29 & 74.29±1.69 & 55.78±0.98 & 49.50±1.01 \\
\cline{2-6}\\
                         & ERM      &78.41±0.14 &73.71±0.17 & \textbf{67.06±0.10} &62.79±0.14\\
                         & D-BAT          & 79.13±0.02 & 76.13±0.03 & 63.22±0.03 & 57.95±0.04 \\
                         & AdvStyle       & 74.45±0.04 & 77.23±1.69 & 61.32±0.34 & 55.31±0.34 \\
                         & EoA            & 72.62±0.32 & 54.67±2.55 & 51.65±1.70 & 49.14±2.60 \\
                         & Mixup          & 50.00±0.00 & 78.74±0.17 & 63.37±0.25 & 56.82±0.26 \\
                         & NCDG           & 76.91±0.13 & 75.09±0.79 & 58.67±0.16 & \textbf{63.41±0.93} \\
                         & SAM           & 50.00±0.00 & 79.70±1.64 & 61.14±0.32 & 58.71±0.23 \\
                         & UDIM           & 54.50±4.18 & 79.41±1.07 & 61.05±0.32 & 58.87±0.24 \\
\cline{2-6} \vspace{-6pt} \\
                         & DBAT PL Ens      & 84.89±0.05 & 76.32±0.03 & 63.16±0.10 & 59.23±0.04 \\
                         & XGB PL Ens      & 84.88±0.19 & \textbf{83.53±0.03} & 63.18±0.03 & 62.53±0.82 \\
                         & \textbf{EXPLOR} (D-BAT)      & 84.36±0.23 & 76.27±0.01 & 63.51±0.16 & 59.01±0.07 \\
                         & \textbf{EXPLOR} (XGB)          & \textbf{87.95±0.25} & 83.11±0.04 & 63.65±0.07 & 61.42±0.35 \\

\bottomrule
\end{tabular}
\normalsize
\label{tab:tableshift}
\end{table}

\subsection{Diversity of Predictions}

 In drug discovery applications, models should predict on structurally diverse compounds. To assess the diversity of model behavior in high confidence out-of-distribution (OOD) predictions, we examine the average variance of fingerprint features for instances with predicted confidence greater than 0.9 on the 3 ChEMBL datasets (var@p\textgreater{}0.9). Higher variance reflects greater heterogeneity among the selected molecules. We observe the following var@p\textgreater{}0.9 on ChEMBL datasets: EXPLOR (\textbf{0.391}), D-BAT (0.337), EoA (0.301), AdvStyle (0.349), Mixup (0.370), and NCDG (0.239). That is, EXPLOR is assigning confident predictions to structurally diverse compounds rather than overfitting to a narrow subset of the chemical space.
\subsection{Full Experiment Results on Tableshift}
In Table~\ref{tab:tableshift}, we report the full results on full results on Tableshift \citep{gardner2023tableshift} datasets.

\subsection{pseudo-labelers Ablations} \label{appx:base_exp}



\begin{table}
\centering
\caption{ChEMBL Datasets' Mean AUPRC Ablating Type of pseudo-labelers.}
\label{tab:base_exp} 
\
\tiny
\begin{tabular}{llll}
\toprule
AUPRC@R\textless{}0.2 & XGBoost & Random Forest & Decision Tree \\
\midrule
PL Ens             & 95.87   & 95.72         & 92.18         \\
EXPLOR                  & 96.64   & 96.15         & 94.15       \\
\bottomrule
\end{tabular}
\end{table}
In this section, We ablate the kind of pseudo-labelers as EXPLOR relies on pseudo-labels during training. reports the mean AUPRC@R$<$0.2 across ChEMBL datasets when using XGBoost, Random Forest (with 100 estimators), and Decision Tree as pseudo-labelers. Performance was comparable between XGBoost and Random Forest, indicating robustness to the choice of strong ensemble models. In contrast, using a weaker pseudo-labelers such as a single Decision Tree led to a performance drop. Nevertheless, EXPLOR consistently outperformed its respective pseudo-labelers, demonstrating its ability to enhance predictions regardless of pseudo-labeler strength (Tab.~\ref{tab:base_exp}. 


\subsection{Per-head Matching Ablation Details}\label{sec:ablation_loss}
In this section we provide details on the per-head matching ablations where we ablate the matching loss scheme on pseudo-labelers and explore a mean-only matching approach on expanded points as an alternative. First, we consider 
utilizing a single-headed (SH) MLP, $f(x)$ ($512 \rightarrow 512 \rightarrow 1$), which is trained via a mean matching loss $
  \mathcal{L}_\mathrm{MM}(f, \{g_j\}_{j=1}^{K}; \mathcal{S}) \equiv \frac{1}{|\mathcal{S}|} \sum_{x \in S} \ell( f(x), \frac{1}{K} \sum_{j =1}^K g_j(x))
$, rather than the per-expert matching loss, $\mathcal{L}_\mathrm{match}$ \eqref{eq:mean_match}
We also explored the effect of training our multi-headed (MH) architecture ($512 \rightarrow 512 \rightarrow 1024$) using only mean-matching (without per-head matching), $
  \mathcal{L}^\prime_\mathrm{MM}(\{h_j\}_{j=1}^{K}, \{g_j\}_{j=1}^{K}; \mathcal{S}) \equiv \frac{1}{|\mathcal{S}|} \sum_{x \in S} \ell(  \frac{1}{K} \sum_{j =1}^K \sigma(h_j(x)), \frac{1}{K} \sum_{j =1}^K g_j(x)).
$ 
\subsection{Bottleneck Ablations Details}\label{sec:bottle_neck}
\begin{table}[hbt!]
\caption{ChEMBL datasets' mean AUPRC ablating hidden layer and output sizes.}
\label{tab:tiny_full} 
\centering
\tiny
\begin{tabular}{lll}
\toprule
& @R$<$.2 & @R$<$1  \\
\midrule                    
Full &  97.12 & 77.28 \\
Full (ML) &  96.94  & 77.25 \\
Full (ERM) & 92.65  & 72.61 \\
Tiny & 96.29  & 76.63 \\
Tiny (ML) & 96.33  & 76.17 \\
Tiny (ERM) & 88.28  & 70.87 \\
\bottomrule
\end{tabular}
\end{table}
In this section, we provide the details and results for the bottleneck ablations. Results are shown in Tab.~\ref{tab:tiny_full}, here `Full' denotes our original $2\times 512$ hidden layer architecture, where `tiny' denotes a $2\times 32$ hidden layer architecture (a $\times16$ decrease in parameters). Moreover, we also consider mean logits `ML,' a final averaging over the multi-head logits that produces a single output unit (i.e., averaging the output weights/bias after training to construct the mean logits network). The performance gap is marginal between the `Full' and `Tiny' model (a $0.85\%$ difference) when using our proposed loss. In contrast, when using empirical risk minimization, we see a $4.87$ times bigger drop in performance between `Full' and `Tiny' models.
This suggests that the bottlenecking properties of our method are key to EXPLOR's performance. Moreover, the results show promise for EXPLOR in resource-constrained settings (such as in IoT applications).

\subsubsection{\textbf{Ablating Bottleneck}}

\begin{table}[hbt!]
\caption{ChEMBL mean AUPRC ablating architecture size.}
\label{tab:tiny_full}
\centering
\scriptsize
\begin{tabular}{lcccccc}
\toprule
& Full & Full (ML) & Full (ERM) & Tiny & Tiny (ML) & Tiny (ERM) \\
\midrule
@R$<$.2 & 97.12 & 96.94 & 92.65 & 96.29 & 96.33 & 88.28 \\
\bottomrule
\end{tabular}
\end{table}
We motivate EXPLOR in terms of how each head is trained to mimic a different pseudo-labeler's predictions on both observed and extrapolated compounds, which encourages the model to learn transferable representations rather than ones that overfit to specific patterns. 
We test this motivation by comparing the original architecture (Full: $2\times 512$ hidden layer) with a stronger bottleneck still, a small architecture (Tiny: $2\times 32$ hidden layer). In addition, we also consider mean logits `ML,' a final averaging over the multi-head logits that produces a single output unit.
 The performance gap is marginal between the `Full' and `Tiny' model (a \emph{0.83\% difference}) when using our proposed loss on ChEMBL dataset. In contrast, when using empirical risk minimization, we observe a $4.37\%$ performance drop on the same datasets. Notably, the 'Tiny' model also speeds up the training time by roughly 50\%. This is particularly attractive in drug discovery pipelines where lightweight models are preferred for large-scale library enumeration and screening, where one may need to score millions of candidate compounds efficiently.

\section{Additional Figures}

Predicted probabilities from EXPLOR network and
pseudo-labelers. We highlight example instances where the
pseudo-labelers initially makes incorrect predictions but are
corrected when we average the predicted probabilities from
EXPLOR network and EXPLOR base.

\end{document}